\documentclass{article}

\usepackage{microtype}
\usepackage{graphicx}
\usepackage{subfigure}
\usepackage{booktabs} 

\usepackage{hyperref}

\usepackage[accepted]{icml2023}

\usepackage{amsmath}
\usepackage{amssymb}
\usepackage{mathtools}
\usepackage{amsthm}

\usepackage[capitalize,noabbrev]{cleveref}

\theoremstyle{plain}
\newtheorem{theorem}{Theorem}[section]

\theoremstyle{definition}
\newtheorem{definition}[theorem]{Definition}

\theoremstyle{remark}

\newcommand{\method}{\textit{SAS}}

\newcommand{\norm}[1]{ \left\| #1 \right\| }

\DeclareMathOperator*{\argmax}{arg\,max}
\DeclareMathOperator*{\argmin}{arg\,min}
\DeclareMathOperator*{\E}{\mathbb{E}}

\newcommand{\LL}{\mathcal{L}}
\newcommand{\x}{\pmb{x}}
\newcommand{\W}{\pmb{W}}

\newcommand{\X}{\pmb{X}}
\newcommand{\muu}{\pmb{\mu}}

\newcommand{\rf}[1]{f(#1)}

\allowdisplaybreaks

\usepackage[textsize=tiny]{todonotes}

\icmltitlerunning{Data-Efficient Contrastive Self-supervised Learning: 
Most Beneficial Examples  
for Supervised Learning Contribute the Least}

\begin{document}

\twocolumn[
\icmltitle{Data-Efficient Contrastive Self-supervised Learning: \\
Most Beneficial Examples  
for Supervised Learning Contribute the Least
}



\icmlsetsymbol{equal}{*}
\begin{icmlauthorlist}
\icmlauthor{Siddharth Joshi}{ucla}
\icmlauthor{Baharan Mirzasoleiman}{ucla}
\end{icmlauthorlist}

\icmlaffiliation{ucla}{Department of Computer Science, University of California Los Angeles, CA 90095, USA.}

\icmlcorrespondingauthor{Siddharth Joshi}{sjoshi804@cs.ucla.edu}

\icmlkeywords{Machine Learning, ICML}

\vskip 0.3in
]



\printAffiliationsAndNotice{} 

\begin{abstract}
Self-supervised learning (SSL) learns high-quality representations from large pools of unlabeled training data. As datasets grow larger, it becomes crucial to identify the examples that contribute the most to learning such representations. This enables efficient SSL by reducing the volume of data required. 
Nevertheless, quantifying the value of examples for SSL has remained an open question. 
In this work, we address this problem for the first time, by proving that examples that contribute the most to contrastive SSL are those 
that have the most similar augmentations to other examples, in expectation. We provide rigorous guarantees for the generalization performance of contrastive learning on such subsets. 
Through extensive experiments, we show that we can safely exclude 20\% of examples from CIFAR100 and 40\% from STL10 and TinyImageNet, without affecting downstream task performance. In general, subsets selected by our method outperform random subsets by over 3\% across these datasets. Interestingly, we also discover the subsets that contribute the most to contrastive learning are those that contribute the least to supervised learning. Code available at https://github.com/bigml-cs-ucla/sas-data-efficient-contrastive-learning.
\vspace{-2mm}
\end{abstract}
\vspace{-6mm}
\section{Introduction}
Large datasets power modern machine learning models. 
However, a key question is: what data points are essential for learning and whether more data will always yield better performance?
Answering this question is crucial as it can reduce the substantial costs of training on large datasets, boost  performance of the trained models and guide data collection.
This has motivated a body of recent research on finding the most essential subsets for supervised learning \cite{dp_toneva_forgettability_2019,dp_el2n_paul_deep_2021,dp-craig-mirzasoleiman20a,dp_mindermann_prioritized_2022,dp_sorscher_beyond_2022,dp_swayamdipta_dataset_2020}.
However, as datasets grow larger, obtaining high-quality labels for them 
becomes prohibitively expensive. 
As a result, there has been 
a surge in self-supervised (SSL) pretraining on large un-labeled dataset \cite{cl_simclr,grill2020bootstrap,chen2021exploring,zbontar2021barlow}. Nevertheless, finding the most important data points for SSL has remained an open question.

Finding the examples that contribute the most to SSL is indeed very challenging. When labels are available, the value of every example for learning can be quantified based on its loss 
(or confidence of the prediction) 
or gradient norm. 
Effectively, difficult-to-learn examples i.e. those with high loss or large gradient norm during training are the ones that contribute the most to minimizing the training loss. 
However, in the absence of labels, SSL methods cluster examples based on their similarity to the other data points. Therefore, 
the SSL loss and gradient of every example is tightly coupled with that of the other examples in the dataset. 
Hence, dropping an example affects the loss and gradient of all the other examples.
This makes data selection inherently more challenging for SSL as compared to supervised learning.

In this work, we address the above challenge for the first time and find examples that provably contribute the most to SSL. In particular, we focus on \textit{contrastive} SSL which 
learns representations by maximizing the alignment between augmented views of the same examples and minimizing the similarity between augmented views of different examples \cite{cl_simclr,zbontar2021barlow,oord2018representation}. 
We prove that examples that contribute the most to contrastive learning are those that have the highest expected similarity between their augmented views and the augmented views of other examples in their latent class. 
Effectively, such examples pull different groups of examples in a class together and enable the contrastive loss to maximally push away representations of examples in different classes.
We show that such examples 
(1) ensure a high alignment between augmented views of examples in every class, and (2) preserve the centers of class representations learned by contrastive learning on the full data.
We leverage the above properties to
provide a generalization guarantee for a linear classifier trained on the 
representations obtained by applying contrastive learning to the subset.

We observe that, perhaps surprisingly, examples that contribute the most to contrastive learning contribute the least to supervised learning. In particular, we quantify the difficulty of examples for supervised learning using confidence of the predictions as well as the forgetting score \cite{dp_toneva_forgettability_2019}, i.e. the number of times an examples is misclassified after being correctly classified during the training. We show that examples that contribute the most to contrastive learning are the easy examples with a high confidence and low forgetting score for supervised learning. Such examples can be safely excluded from a supervised learning pipeline, without harming the accuracy \cite{dp_toneva_forgettability_2019}.  
In contrast, difficult-to-learn examples that contribute the most to supervised learning can significantly hurt contrastive learning performance.


We extensively evaluate the performance of our method, \method, which selects \textbf{S}ubsets that maximize \textbf{A}ugmentation \textbf{S}imilarity to the full data, on various datasets and using different contrastive learning methods.
We first apply \method\ to CIFAR10, CIFAR100 \cite{krizhevsky2009learning}, STL10 \cite{coates2011analysis} and TinyImageNet \cite{tiny_imagenet}, with ResNet50 using SimCLR \cite{cl_simclr}.
We show that using \method, 
up to 20\% of examples from CIFAR100 and 40\% from STL10 and TinyImageNet \cite{imagenet}, can be safely excluded without harming the downstream performance. Similarly, for BYOL, using \method\ to discard 20\% of examples from STL10 can even outperform downstream performance of the full data by 2\%. In general, \method\ subsets outperform random subsets by over 3\% across these datasets and methods including SimSiam \cite{chen2021exploring}, MoCo \cite{cl_he_momentum_2020} and BYOL \cite{grill2020bootstrap}. 
We also demonstrate that the subsets that contribute the most to SSL can be efficiently extracted can be efficiently extracted early-in-training or using a smaller proxy model.



\section{Related Work}
\textbf{Contrastive Learning.}
Contrastive learning has recently emerged as a performant self-supervised framework to learn representations that capture semantically relevant information from the data. The key idea behind this family of algorithms is learning representations by maximizing agreement between augmented views of the same example (positive pairs) and minimizing agreement between augmented views of different examples (negative pairs) 
\cite{cl_simclr,zbontar2021barlow,grill2020bootstrap,chen2021exploring,cl_he_momentum_2020}. 
To improve the performance of contrastive learning, re-weighting the negative pairs in the contrastive loss \cite{cl_chuang_debiased_2020} or re-weighting the loss to emphasize the hard negatives \cite{cl_robinson_hard_neg_2020} has been recently explored.
Here, we aim to find subsets of examples that contribute the most to contrastive learning. The above reweighting
strategies are orthogonal to our work and can be applied to the subsets found by our method.

\textbf{Contrastive Learning Theory.}
A recent line of theoretical works has studied contrastive learning.
In particular, under conditional independence between positive pairs given the label, representations learned
by contrastive learning algorithms can achieve small errors in the downstream linear classification task \cite{arora2019theoretical, cl_theory_saunshi_2019, tosh2021contrastive}. The independence assumption was relaxed by \cite{cl_provable_guarantees}, which showed that minimizing spectral-based contrastive loss results in spectral clustering on the augmented distribution and provides generalization guarantee for linear evaluation. \citet{wang2020understanding} proved that asymptotically, the contrastive loss optimizes alignment (similarity) of positive pairs and uniformity of the representations 
on the hypersphere, relating them to positive effects on downstream tasks. 
The recent result of \cite{huang2021towards} showed that contrastive learning using the more general InfoNCE \cite{oord2018representation} or cross-correlation loss \cite{zbontar2021barlow} maximizes alignment of positive pairs as well as divergence of centers of the latent class representations. 
Here, we build on this work and show that subsets that contribute the most to contrastive learning introduce minimal error on the alignment and divergence of centers of class representations learned on the full data. Leveraging the above properties, we provide generalization guarantees for downstream performance of representations learned on such subsets. \looseness=-1

\textbf{Essential Subsets for Supervised Learning.}
There has been a recent body of efforts on finding the most important subsets for supervised learning. Empirical methods commonly rank examples from easiest to hardest---based on  confidence, loss or gradient---and curate subsets preserving the hardest examples.
\citet{coleman2020selection} used a smaller trained proxy model to find the most uncertain examples to train a larger model.
\citet{dp_toneva_forgettability_2019} 
selects examples with highest forgetting score, i.e., the number of times they transition from being classified correctly to incorrectly during training. \citet{dp_swayamdipta_dataset_2020} selects examples with the highest variance of predictions during training.
\citet{dp_el2n_paul_deep_2021} selects examples with the lowest expected gradient norm over multiple initializations. More theoretically motivated approaches iteratively select subsets by importance sampling based on gradient norm \cite{katharopoulos2018not} or select weighted subset of examples which closely capture the full gradient \cite{dp-craig-mirzasoleiman20a,pooladzandi2022adaptive,killamsetty2021grad}.

In contrast, we show, for the first time, that easy-to-learn examples with highest confidence and lowest forgetting score that contribute the least to supervised learning are the most beneficial for unsupervised contrastive learning. \vspace{-3mm}

\section{Problem Formulation}
Assume we have a dataset $\X\!=\!\{\x_i\}_{i\in V}$ of $n\!=\!|V|$ training examples drawn i.i.d. from an unknown distribution. Each example belongs to one of the $K$ latent classes i.e. $V=\{V_1 \cup \cdots \cup V_K\}$, 
but the 
corresponding class labels are not known at training time.

Contrastive Learning learns representations of examples in the training data, by learning an encoder $f$ that maximizes agreement between representations of differently augmented views of the same example (i.e. positive pairs) and minimizes agreement between representations of augmented views of different examples (i.e. negative pairs). This is achieved by minimizing the following InfoNCE loss \cite{oord2018representation}:
\begin{equation}\label{eq:cl_loss}
\LL_{cl}(V)\!=\!-\!\!\!\!\E_{i,j\in V}\E_{\substack{\x_1,\x_2\in A(\x_i)\\ \x^-\!\in A(\x_j)}} \!\!\!\log\! \frac{e^{f(\x_1)^T f(\x_2)}}{e^{f(\x_1)^T f(\x_2)}\!+\!e^{f(\x_1)^T f(\x^-)}},
\end{equation}
where $A(\x)$ is the set of augmented views of example $\x$.

The performance of contrastive learning is evaluated by training a linear classifier on the learned representation using labels:
\begin{equation}
    g_f^l(\x)=\argmax_{k\in[K]} (\W f(\x)+\pmb{b})_k
\end{equation}
However, to simplify the theoretical analysis, we follow \cite{huang2021towards} and consider a non-parametric nearest neighbor (NN) classifier:
\begin{equation}
    g_f(\x)=\argmin_{k\in[K]}\|f(\x)-\muu_k\|,
\end{equation}
where $\muu_k \!\!:=\! \E_{i\in V_k} \!\E_{\x'\in A(\x_i)}[f(\x')]$ is the center of class $V_k$. \looseness=-1

The linear classifier learned on the labels $g_f^l$ is guaranteed to perform at least as well as the NN classifier $g_f$ \cite{huang2021towards}.
Therefore, we use the classification error rate of the NN classifier to bound the worst-case performance of the linear classifier: \vspace{-1mm}
\begin{equation}
    \xi(g_f(V)) = \sum_{k=1}^K \mathbb{P}[g_f(\x_i) \neq k,\forall i \in V_k].
\end{equation}
We note that in our experiments, we evaluate our method using the downstream accuracy of the \textit{linear classifier}, and our theoretical guarantees on the NN classifier also upper-bound the error of the linear classifier. 

Our goal is to find a subset  $S\subseteq V$ of at most $r$ training examples, such that the encoder $f^S=\argmin_{f}\LL_{cl}(S)$ obtained by minimizing the contrastive loss on the subset, allows the NN classifier to obtain a similar error on the \textit{full data}. Formally, we aim to solve the following problem:
\begin{equation}
    S^*=\argmin_{\substack{S\subseteq V,|S|\leq r}} [|\xi(g_{f^S}(V)) - \xi(g_{f}(V))|].
\end{equation}
\section{The Most Important Subsets for SSL}
\label{sec:method}

\begin{figure*}
    \centering
    \subfigure{\includegraphics[width=0.4\linewidth, keepaspectratio]{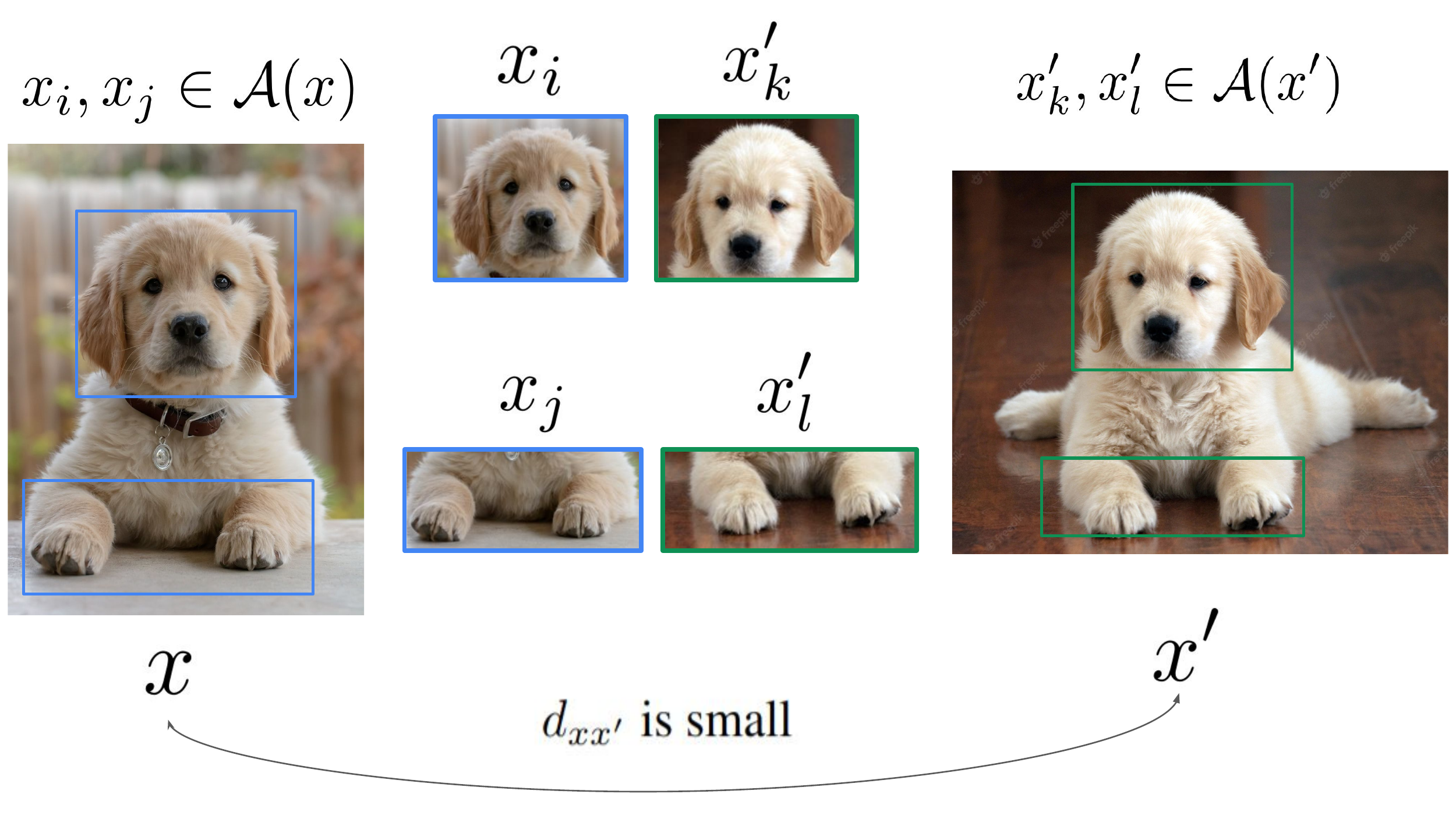}}
    \hspace{8mm}
    \subfigure{\includegraphics[width=0.4\linewidth, keepaspectratio]{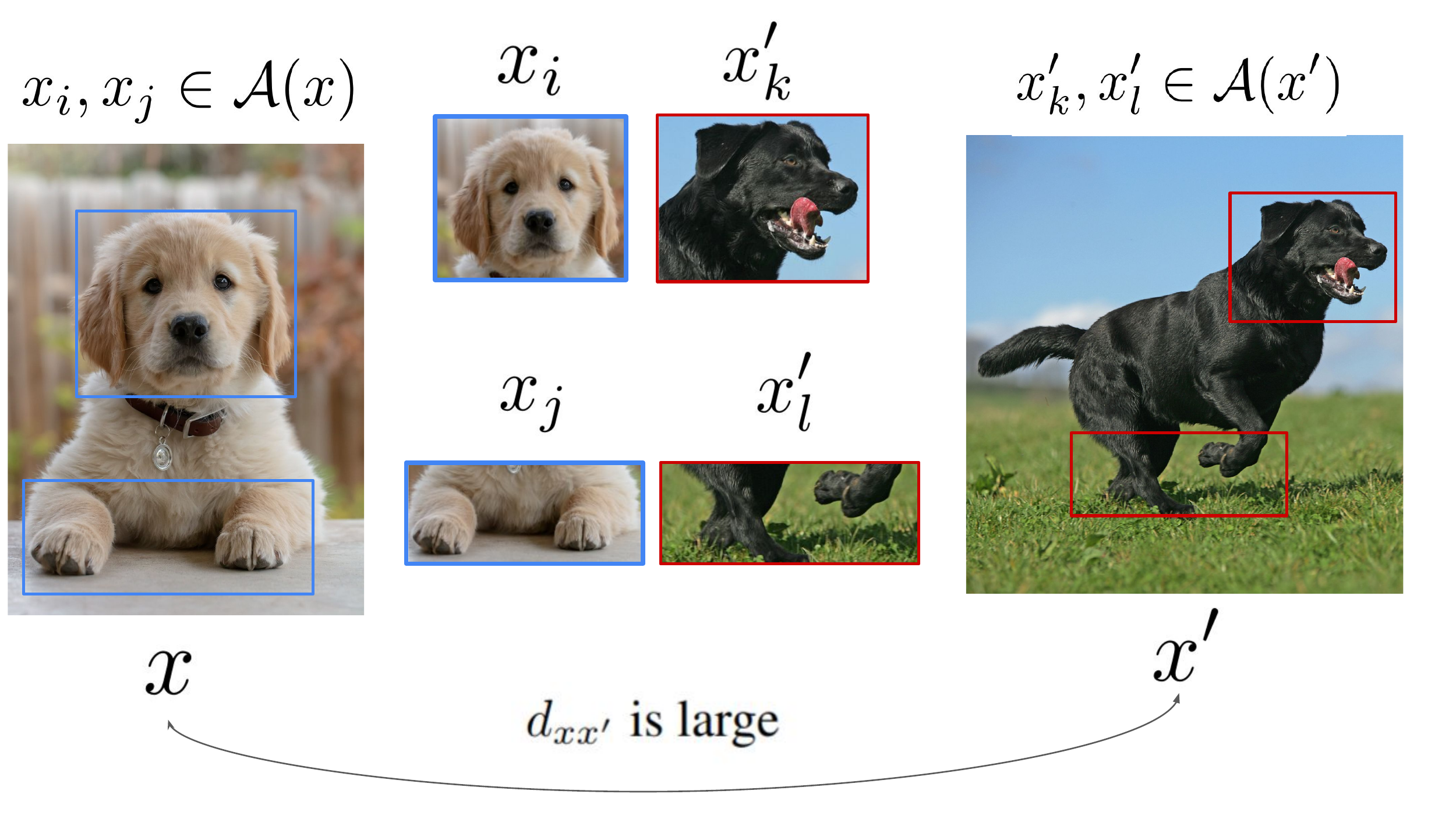}}
    \vspace{-4mm}
    \caption{Visualizing Expected Augmentation Distance $d_{x, x'}$. Pair of examples on left shows two examples that are semantically very similar as seen by their augmentations being very similar to each other, thus the expected augmentation distance between them is small. In contrast, pair of examples on the right are not as semantically similar, thus have augmentations that are very dissimilar to each other.}
    \label{fig:exp_aug_dist}
    \vspace{-4mm}
\end{figure*}


We start by investigating which properties the subset $S^*$ must satisfy, such that the learned representations on the subset provide small downstream classification error. To do so, we rely on recent theoretical results on optimization and generalization of contrastive learning. 
In particular, the recent results of \citet{huang2021towards} showed that the generalization performance of representations
obtained with contrastive learning 
dependents on: (1) alignment of positive pairs, (2) divergence of class centers and (3) concentration of the augmented data.
Alignment 
captures the similarity between representations of augmented views of examples, in expectation. Good alignment requires 
all augmented views of an example to have similar representations.
Divergence of class centers captures how distant class centers $\muu_l$ and $\muu_k$ are. Good divergence results in large enough distance between all pairs of class centers, i.e., small $\muu_l^T\muu_k ~\forall l,k \in [K]$.
Concentration of augmented data 
is determined by the data distribution and augmentation pipeline. Specifically,
let $V^0_k \subseteq V_k$ be the subset of examples in every class $k \in K$ that 
share at least one very similar augmented view: 
$\sup_{\x_1,\x_2\in V_k} \min_{\x'_1\in A(\x_1),\x'_2\in A(\x_2)}\!\norm{\x_1-\x_2}\!\leq\! \delta$ for small $\delta>0$. If for every latent class $k \in [K]$, $V^0_k$ is large enough ($|V_k^0|\geq \sigma |V_k|$ for large $\sigma\in (0,1]$), then the classes have sharp concentration of augmented data.
In this case, good alignment and divergence guarantee good generalization performance for the downstream NN classifier.

While concentration of the augmentations is 
independent of the contrastive loss,
minimizing the contrastive loss effectively aligns the augmented views of the examples and results in good divergence of the class centers.
Formally, for a normalized encoder $\norm{f}\!=\!1$, the InfoNCE loss in Eq. \eqref{eq:cl_loss} can be written as:\looseness=-1
\begin{align}\label{eq:loss_equiv}
    \LL_{cl}&(V)
    \!=\!\frac{1}{2}[
    \underbrace{\E_{i\in V}\E_{\x_1,\x_2\in A(\x_i)}\!\norm{f(\x_1)\!-\!f(\x_2)}^2]}_{\LL_{align}(V) \text{: Related to Alignment}}\!-\!1\\
    &+\!\!\!\!\E_{i,j\in V} \;\E_{\substack{\x_1,\x_2\in A(\x_i)\\ \x^-\!\in A(\x_j)}}\log \big(e^{f(\x_1)^Tf(\x_2)}+ \underbrace{e^{f(\x_1)^Tf(\x^-)}}_{\text{Related to Divergence}} \!\!\!\big).\nonumber
\end{align}
The first term in the RHS of Eq. \eqref{eq:loss_equiv} is closely related to the alignment and the second term in the RHS is related to the divergence of class centers. 

\textbf{Alignment. }
Minimizing the first term in the RHS of Eq. \eqref{eq:loss_equiv} aligns augmented views of the training examples in expectation, and results in a small probability $R_{\epsilon}(V)$ for examples to still have non-aligned augmented views, i.e, the largest distance between their augmented views is larger than $\epsilon$: \looseness=-1
\begin{equation}\label{eq:R}
    R_{\epsilon}(V) = \mathbb{P} \Big[ i\in V: \!\!\!\!\!\!\sup_{\x_1,\x_2\in A(\x_i)} \!\!\!\!\|f(\x_1)-f(\x_2)\|>\epsilon \Big]. 
\end{equation}
In particular, for a $L$-Lipschitz continuous encoder $f$, we have that \cite{huang2021towards}: 
\begin{align}\label{eq:R_upper} \vspace{-3mm}
    R_{\epsilon}(V)\leq \eta(\epsilon)\!\cdot\!\sqrt{\LL_{align}(V)},
\end{align}
where $\!\LL_{align}(V)\!=\!\E_{i\in V}\! \E_{\x_1,\x_2\in A(\x_i)}\!\|f(\x_1)\!-\!f(\x_2)\|^2$ is the alignment loss; $\!\eta(\epsilon)\!=\!\mathcal{O}(\frac{1}{\epsilon})\!$ is a function of $\epsilon$ and the transformations used for data augmentations. 

\textbf{Divergence.} Minimizing the second term in RHS of Eq. \eqref{eq:loss_equiv} pushes away the class centers, i.e., expected representation of examples in a class, and yields a small $\muu_k^T\muu_l$ for all $k,l\in[K]$. Effectively, it maximizes the distance between different class representations.

Minimizing the InfoNCE loss in Eq. \eqref{eq:loss_equiv} minimizes both terms in the RHS, thus ensuring good alignment and divergence.
With good alignment (small $R_\epsilon$) and good divergence (small $\muu_k^T \muu_l$), the NN classifier $g_f$ can correctly classify all the examples in the main part of every class that have concentrated and aligned augmented views. If the majority of examples in every class have a high  concentration of augmented data is large (large $\sigma$), good generalization is guaranteed.
Formally, 
\begin{theorem}[\citealt{huang2021towards}]
\label{thm:huang_main_paper}
For any $l, k \in [K]$, if 
\begin{equation}
{\muu_k}^T{\muu_l} < \phi (\sigma, \delta, \epsilon),
\end{equation}
then the downstream error rate of NN classifier is
\begin{align}
\xi(g_{f}(V))\leq (1-\sigma)+R_{\epsilon}(V).
\end{align}
\end{theorem}
Exact form of {$\phi(\sigma, \delta, \epsilon)$} is discussed in Appendix \ref{appendix:proof}.

\subsection{Subsets that Preserve Alignment and Divergence}

We rely on the above observations to find a subset that, when used to learn representations, provides similar generalization performance to that of the full data, for the downstream NN classifier.
The key idea of our approach is to find a subset, such that minimizing the contrastive loss on this subset: (1) results in good alignment for all the examples, and (2) preserves the class centers of full data. In doing so, we ensure that the divergence of the class centers is preserved. 
If such a subset can be found, minimizing the contrastive loss in Eq. \eqref{eq:cl_loss} on the subset results in good alignment and divergence on the full data, hence guarantees similar generalization performance for the downstream NN classifier. 


Next, we introduce the notion of expected augmentation distance and discuss how it can be leveraged to find a subset that satisfies the above two conditions. 

We start by defining the expected augmentation distance:
\begin{definition}[\textbf{Expected augmentation distance}]
We define the expected augmentation distance between examples $i,j\in V$ as the expected $l_2$ norm between all pairs $(\x,\x')$ of augmented examples, such that $\x\in A(\x_i)$ and $\x'\in A(\x_j)$. Formally, for every pair of examples $i,j\in V$ we have: 
\begin{equation}\label{eq:dist}
    d_{i,j}=\E_{\substack{\x\in A(\x_i),\x'\in A(\x_j)}} \|\x - \x'\|.
\end{equation}
\vspace{-7mm}
\end{definition}
Intuitively, expected augmentation distance captures the \textit{semantic dissimilarity} between every pair of examples. That is, two examples that are semantically similar have a small expected augmentation distance. We visualize examples with small and large expected augmentation distance in Fig. \ref{fig:exp_aug_dist}. \looseness=-1



\subsection{Ensuring Good Alignment}
\label{sec:good_alignment}
First, we address finding a subset that, when used to minimize the contrastive loss, aligns the augmented views of all the training examples. 
From Eq. \eqref{eq:R_upper}, we know that minimizing the alignment loss $\LL_{align}$, directly minimizes the probability $R_{\epsilon}(V) $ of examples with non-aligned augmented views. That is   $R_{\epsilon}(V)\leq \eta(\epsilon)\!\cdot\!\sqrt{\LL_{align}(V)}$.
Here, we find a subset $S_k\subseteq V_k$ of examples from every latent class $k$ that ensures small $R_{\epsilon}(V_k)$,  i.e., probability that examples in $V_k$ are not well-aligned. For every (arbitrary) subset $S_k\subseteq V_k$ of size $r_k=|S_k|$ selected from class $k$ with $n_k=|V_k|$ examples, we can upper-bound the probability $R_{\epsilon}(V_k)$ based on the alignment loss of the subset i.e. 
$\LL_{align}(S_k)$. 
In particular, using $R_{\epsilon}(V_k) \leq \eta(\epsilon) \!\cdot\! \E_{i\in V_k}\E_{\x_1,\x_2\in A(\x_i)} \norm{f(\x_1)-f(\x_2)} \leq \eta(\epsilon) \sqrt{\LL_{align}(V)}$ 
\cite{huang2021towards}, we can write:
\begin{align}
    &\hspace{-2mm}R_{\epsilon}(V_k) \nonumber \\
    &\leq \eta(\epsilon) \!\cdot\! \E_{i\in V_k}\E_{\x_1,\x_2\in A(\x_i)} \norm{f(\x_1)-f(\x_2)} \\
    &\!= \frac{\eta(\epsilon)}{n_k} \!\cdot\! \Bigg( \sum_{i \in S_k}\E_{\x_1,\x_2\in A(\x_i)} \norm{f(\x_1)-f(\x_2)}\nonumber \\
    &\hspace{1.4cm} + \!\!\!\sum_{i \in V_k\setminus S_k}\E_{\x_1,\x_2\in A(\x_i)} \norm{f(\x_1)-f(\x_2)} \!\Bigg) \label{eq:align_bound_1}\\
    &\!\leq \frac{\eta(\epsilon)}{n_k} \!\cdot\! \Bigg( \sum_{i \in S_k}\E_{\x_1,\x_2\in A(\x_i)} \norm{f(\x_1)-f(\x_2)} \nonumber \\
    &\hspace{.4cm} + \!\!\!\sum_{i \in V_k\setminus S_k}\!\!\![2 \min_{j\in S_k}\E_{\substack{\x_1\in A(\x_i),\\\x_2\in A(\x_j)}}\!\! \norm{f(\x_1)\!-\!f(\x_2)}]\!\Bigg), \label{eq:align_bound_2}
\end{align}    
Detailed steps of getting Eq. \eqref{eq:align_bound_2} from Eq. \eqref{eq:align_bound_1}  can be found in the Appendix \ref{appendix:detailed_steps}. Note that the first term in Eq. \eqref{eq:align_bound_2} is exactly $\frac{\eta(\epsilon)}{n_k}\sqrt{\LL_{align}(S_k)}$.
Hence, for a $L$-Lipschitz continuous encoder $f$, where $\norm{f(\x)-f(\x')}\leq L \norm{\x-\x'}~\forall \x,\x'$, we have:
\begin{equation}
    R_{\epsilon}(V_k)\leq \frac{\eta(\epsilon)}{n_k} \Bigg(\!{r_k}\sqrt{\LL_{align}(S_k)} + 2L\!\!\!\!\sum_{i\in V_k\setminus S_k}\!\!\! \min_{j\in S_k} d_{i,j}\!\Bigg). \nonumber
\end{equation}
The alignment loss on the subset $\LL_{align}(S_k)$ 
can be effectively minimized by contrastive learning on the subset using the InfoNCE loss. We also empirically show in Appendix \ref{appendix:experiments} Fig. \ref{fig:align} that alignment loss on the 
subsets we find 
for contrastive learning is smaller 
than the alignment loss on the full data, i.e., $\LL_{align}(S_k)\leq \LL_{align}(V_k)$. 
Therefore, training on a subset $S_k\subseteq V_k$ 
introduces at most the following error on $R_{\epsilon}(V_k)$, i.e., the probability for any example in $V_k$ to have a distance larger than $\epsilon$ between its augmented views: \looseness=-1
\begin{equation}\label{eq:r_error}
    \nu_R^k 
    \leq \frac{2L\eta(\epsilon)}{n_k}\!\! \sum_{i\in V_k\setminus S_k}\!\!\! \min_{j\in S_k} d_{i,j},
\end{equation}
Therefore, the subset $S_k\subseteq V_k$ with \textit{smallest expected augmentation distance} $d_{i,j}$ (semantic similarity) to the \textit{rest} of the examples in the class $V_k\setminus S_k$ can best align augmentations of all the examples in the class $V_k$. 

\textbf{Remark.} 
Eq. \eqref{eq:r_error} shows that the subset $S_k$ that aligns augmented views of all the examples in a class $V_k$ should have an element that is sufficiently similar to any other example 
in the class. In other words, the subset should contain examples that are \textit{representative of every group} of examples in the class.
\vspace{-2mm}

\subsection{Preserving the Class Centers}
Next, we discuss finding a subset that captures the center of every latent class $\muu_k$. 

For every (arbitrary) subset $S_k\subseteq V_k$ of size $r_k=|S_k|$ selected from class $k$ with $n_k=|V_k|$ examples, we can write: \vspace{-2mm}
\begin{align}
    \muu_k \!&= \E_{\substack{\i\in V_k, \\ \x'\in A(\x_i)}} [\rf{\x'}] \nonumber\\
    &\!\!\!=\E_{\substack{i\in V_k,\\ \x'\in A(\x_i)}}\!\!\! [\rf{\x'}]\;-\E_{\substack{j\in S_k,\\ \x''\in A(\x_j)}}\!\!\! [\rf{\x''}]\;+\E_{\substack{j\in S_k,\\ \x''\in A(\x_j)}}\!\!\! [\rf{\x''}] \nonumber\\
    &\!\!\!=\frac{1}{n_k}\!\sum_{i\in V_k} \!\!\E_{\x'\in A(\x_i)} [\rf{\x'}] \!-\! \frac{1}{r_k}\!\sum_{j\in S_k} \!\!\E_{\x''\in A(\x_j)} [\rf{\x''}] + \muu_k^S \nonumber\\
    &\!\!\!=\!\!\frac{1}{n_k\!\cdot \!r_k} \Big[r_k\!\!\sum_{i\in V_k}\!\!\E_{\x'\!\in\! A(\x_i)}\!\!\!\! [\rf{\x'}] \!-\! n_k\!\!\sum_{j\in S_k}\!\!\E_{\x''\!\in \!A(\x_j)}\!\!\!\! [\rf{\x''}]\Big]\!+\!\muu_k^S \nonumber\\
    &\!\!\!=\!\frac{1}{n_k\!\cdot \!r_k} \!\sum_{i\in V_k} \!\sum_{j\in S_k} \!\!\big[\E_{\x'\!\in\! A(\x_i)}[\rf{\x'}]-\!\!\!\!\!\!\E_{\x''\!\in\! A(\x_j)}\![\rf{\x''}]\big]\!+\!\muu_k^S \nonumber\\
    &\!\!\!=\frac{1}{n_k\!\cdot \!r_k} \sum_{i\in V_k} \sum_{j\in S_k} \! \E_{\substack{\x'\in A(\x_i),\\ \x''\in A(\x_j)}}\![\rf{\x'}\!-\!\rf{\x''}] +\muu_k^S 
\end{align}
Hence, for a $L$-Lipschitz continuous encoder $f$, where $\norm{f(\x)-f(\x')}\leq L \norm{\x-\x'}~\forall \x,\x'$, we can upper-bound the normed difference between the center of class $V_k$ and subset $S_k$ as follows:
\begin{align}\label{eq:mu_error}
    \nu_{\mu}^k=\|\muu_k-\muu_k^S\|\!\leq 
    L \cdot \E_{\substack{i\in V_k,\\ j \in S_k}}[d_{i,j}]. 
\end{align}
That is, the subset that preserves the center of class $k$, can be found by \textit{minimizing the expectation of expected augmentation} distances (semantic similarity) between examples in the subset $S_k$ and \textit{all} the data points $V_k$ in class $k$. 

\textbf{Remark.} 
Eq. \eqref{eq:mu_error} implies that a subset $S_k$ that captures the centre of class $k$, should be similar to {all} the examples in the class, in expectation. 
Such a subset contains examples from dense regions with sharp concentration of augmented data. Such examples \textit{best represent the entire class}.
\vspace{-2mm}

\vspace{-1mm}\subsection{Minimizing the Alignment and Divergence Error}\vspace{-2mm}

\begin{figure}    \centering\includegraphics[width=\linewidth, keepaspectratio]{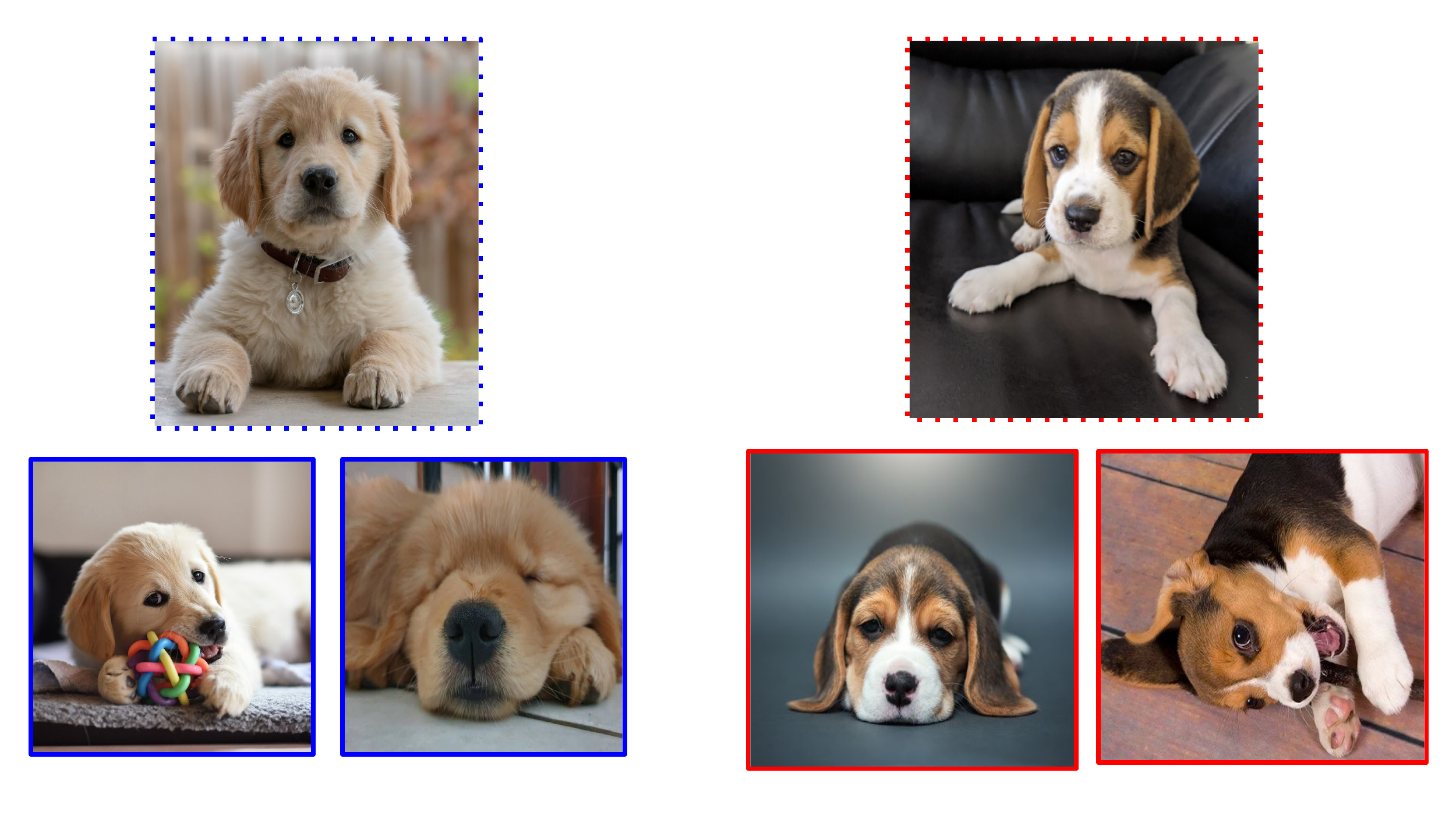}
    \vspace{-8mm}
    \caption{Most representative examples: examples in top row are each representative of their group (e.g. breed) in class \textit{dog}.}
    \label{fig:core_examples}
    \vspace{-4mm}
\end{figure}



Based on Eq. \eqref{eq:r_error} and \eqref{eq:mu_error}, we find
the subset that ensures alignment of all data points in class $k$ and closely captures the center of the class, 
by solving the following 
problem: 
\begin{equation}\label{eq:subset}
    S_k^*=\!\!\!\!\underbrace{\argmin_{\substack{S\subseteq V_k},\\ |S|\leq r_k}\E_{\substack{i\in V_k,\\ j\in S_k}}\![d_{i,j}]}_{\text{Captures class center}} \quad \text{s.t.} \quad \underbrace{\min_{j\in S_k}d_{i,j}\!\leq\! \delta~~ \forall i\!\in\! V_k}_{\text{Ensures alignment}}.
\end{equation}
Problem \eqref{eq:subset} is NP-hard as it involves calculating the value of the objective over an exponential number of subsets.
To efficiently find a subset that captures the class center and contains representatives from different groups of examples in a class, 
we rely on the following objective which minimizes the \textit{sum} of expected augmentation distance between examples in the subset $j\in S_k$ and the {rest of examples} in the class $V_k\setminus S_k$:\looseness=-1 \vspace{-3mm}
\begin{equation}\label{eq:subset_relax}
    S_k^*=\!\!\!\!\argmin_{\substack{S\subseteq V_k},\\ |S|\leq r_k}\!\sum_{\substack{i\in V_k\setminus S_k}}\sum_{j\in S_k}d_{i,j}.
    \vspace{-1mm}
\end{equation} 
By minimizing the sum of distances (dissimilarities) between the subset and the \textit{rest} of examples, Eq. \eqref{eq:subset_relax} finds examples that are similar to many other examples in their class. In doing so, it 
finds a subset that ensure alignment.
At the same time, the selected examples 
are selected from \textit{dense} regions with sharp concentration of augmented data.
Hence, the subset closely preserves the class center. \looseness=-1

The above minimization problem can be turned into maximizing the following non-monotone submodular\footnote{A set function $F:2^V \rightarrow \mathbb{R}^+$ is \textit{submodular} if $F(e|S)=F(S\cup\{e\}) - F(S) \geq F(T\cup\{e\}) - F(T),$ for any $S\subseteq T \subseteq V$ and $e\in V\setminus T$.}
problem:
\vspace{-2mm}\begin{equation}\label{eq:submodular}
    S_k^*=\!\!\!\!\argmax_{\substack{S\subseteq V_k},\\ |S|\leq r_k}\!\sum_{\substack{i\in V_k\setminus S_k}}\sum_{j\in S_k} C-d_{i,j},
\end{equation}
where $C$ is a big constant. $C-d_{i,j}$ captures the \textit{similarity} between $i$ and $j$.  

Thus, we can find a nearly optimal subset using algorithms designed for maximizing non-monotone submodular functions under a cardinality constraint. First, we rely on the greedy algorithm to identify a subset and then refine it by using unconstrained submodular maximization \cite{mirzasoleiman2016fast}. The greedy algorithm commences with an empty set $S_0=\emptyset$, and at each step $t$, it selects an element $e\in V$ that maximizes the marginal utility $F(e|S_{t})=F(S_{t}\cup\{e\}) - F(S_{t})$. Formally, $S_t = S_{t-1}\cup\{{\arg\max}_{e\in V} F(e|S_{t-1})\}$. For unconstrained maximization, we utilize the double-greedy algorithm \cite{buchbinder2015tight}, which initializes $S^\alpha=\emptyset$ and $S^\beta=S_T$, where $S_T$ is the subset found by the final iteration of the greedy algorithm. It then computes $a_e=F(e|S^\alpha)$ and $b_e=F(S^\beta\setminus \{e\})$ for all $e\in V$. Subsequently, it adds examples for which $a_e\geq b_e$ to $S^\alpha$ and removes examples for which $a_e< b_e$ from $S^\beta$, eventually setting $S^\alpha=S^\beta$. The time complexity of the greedy algorithm is $\mathcal{O}(nk)$ to identify $k$ out of $n$ examples, which can be further expedited through lazy evaluation \cite{minoux2005accelerated}. The double-greedy approach applied to the subset has a complexity of $\mathcal{O}(k)$.
Thus, the subset can be found very efficiently. \looseness=-1


\textbf{Remark.} Intuitively, as the subsets selected from different classes have a small expected augmentation distance to 
all the different groups in the class, 
they pull together all the examples in a class during contrastive learning and let the representations of a class to cluster closely. At the same time, as they preserve the class centers, they allow the representations of different classes to be effectively pushed away from each other.
In doing so, 
the subsets
effectively minimize the contrastive loss on the full data. 
Note that as $d_{i,j}$ is a property of the data in the \textit{input space}, the subset found by solving Problem \eqref{eq:submodular} ensures good alignment and divergence \textit{throughout contrastive learning}.

Fig. \ref{fig:core_examples} presents a visualization of examples 
in the dog class.
The examples found by Eq. \eqref{eq:submodular} resemble those in the top row, i.e., they contain the core features of the class (e.g. the head and the paws of the puppies) with minimal noise (e.g. the non-standard poses of the puppies in the bottom row). Due to the standard and clear presentation of the core features of their respective groups, the examples in top row have smaller expected augmentation distance to many examples than examples in the bottom row, where some core features may be occluded (e.g. paws not visible) and/or presented in non-standard ways (e.g. open mouth). 

\vspace{-2mm}
\begin{algorithm}[t]
\caption{\method: finding \textbf{S}ubsets that maximize the expected \textbf{A}ugmentation \textbf{S}imilarity}\label{alg}
\begin{algorithmic}[1]
\STATE \textbf{Input:} Subset size $B$, proxy model $f_p$
\STATE \textbf{Output:} Subset $S$
\STATE $\{V_1, ..., V_K \} \leftarrow  \text{approximate latent classes (Sec. \ref{sec:practice})}$ 
\FORALL{$V_k \in \{V_1, \cdots, V_K \}$}
    \FORALL{$i,j \in {V_k}$}
            \STATE $s_{i,j}= \left< f_p(\x_i), f_p(\x_j)\right>$
    \ENDFOR
    \STATE $S_k \leftarrow \{\}$
    \STATE $r_k \leftarrow \frac{|V_k|}{|V|}\cdot B$
    \STATE $F(S_k)=\sum_{\substack{i\in V_k\setminus S_k}}\sum_{j\in S_k} s_{i,j}$
    \WHILE{$|S_k| \leq r_k$}
    \STATE $e \leftarrow \text{argmax}_{e \in V_k \setminus S_k} F(e|S_{k})
    $
    \STATE $S_{k} \leftarrow S_k \cup \{ e\}$
    \ENDWHILE
    \STATE $S_k \leftarrow \text{double-greedy}(S_k)$
\ENDFOR 
\STATE \textbf{return} $S=\{S_1\cup\cdots \cup S_K\}$
\end{algorithmic}
\end{algorithm}
 



Next, we provide a generalization guarantee for contrastive learning from the subset.
The following theorem shows that if contrastive learning on the subset provides a small extra divergence on the center of examples selected from different classes compared to that of full data,
the downstream NN classifier will have a similar generalization performance to that of contrastive learning from full data.
\begin{theorem}\label{thm:diverge}



Assume f is a normalized encoder and the subset $S_k$ selected by Eq. \eqref{eq:submodular} has $\nu^k_R$ error (Eq. \ref{eq:r_error}) in capturing $R_{\epsilon}(f,V_k)$ and $\nu^k_{\mu}$ error (Eq. \ref{eq:mu_error}) in capturing the center of class $k$. If for any pair of classes $k, l \!\in\![K]$, we have: \looseness=-1 \vspace{-2mm}
\begin{align}
{\muu_k^S}^T{\muu_l^S} < &\phi(\sigma, \delta, \epsilon) -\\
&\big( C \nu^k_R+2(\max\{\nu^k_{\mu},\nu_{\mu}^l\})^2 
+4\max\{\nu^k_{\mu},\nu_{\mu}^l\}) \big).\nonumber
 \end{align}
 %
%
%
where $\phi(\sigma, \delta, \epsilon)$ is the requirement on divergence of full data class centers in Theorem \ref{thm:huang_main_paper} and $C$ is a constant, then the generalization error of the model trained on the subset can be bounded by: \vspace{-1mm}
\begin{align}
    \xi(g_{f^S}(V)) \leq (1 - \sigma) + R_\epsilon + \nu_R.
\end{align}
\end{theorem}
\vspace{-4mm}Theorem \ref{thm:diverge} shows that if the subset captures the class centers and alignment closely (i.e. $\nu_R$ and $\nu_{\mu}$ are small), then minimizing the contrastive loss on the subset 
provides a similar divergence 
to that of full data, 
and thus a similar downstream generalization performance 
for the 
NN classifier is guaranteed.

The proof can be found in Appendix \ref{appendix:proof}, where we also discuss that $C\nu_R$ is generally small.
Fig. \ref{fig:diverge} in Appendix \ref{appendix:experiments} confirms that divergence of \textit{full data} class centers when training on sufficiently large subsets found by Eq. \eqref{eq:submodular} is in fact better than that of training on the full data. This explains the similar or even superior generalization performance of models trained on \method\ subsets to models trained on the full data. \looseness=-1

\vspace{-4mm}
\subsection{SAS: Finding the Subset in Practice}\label{sec:practice}
\vspace{-1mm}
\begin{figure*}[h]
    \centering
    \subfigure[CIFAR10]{\includegraphics[width=0.24\linewidth]{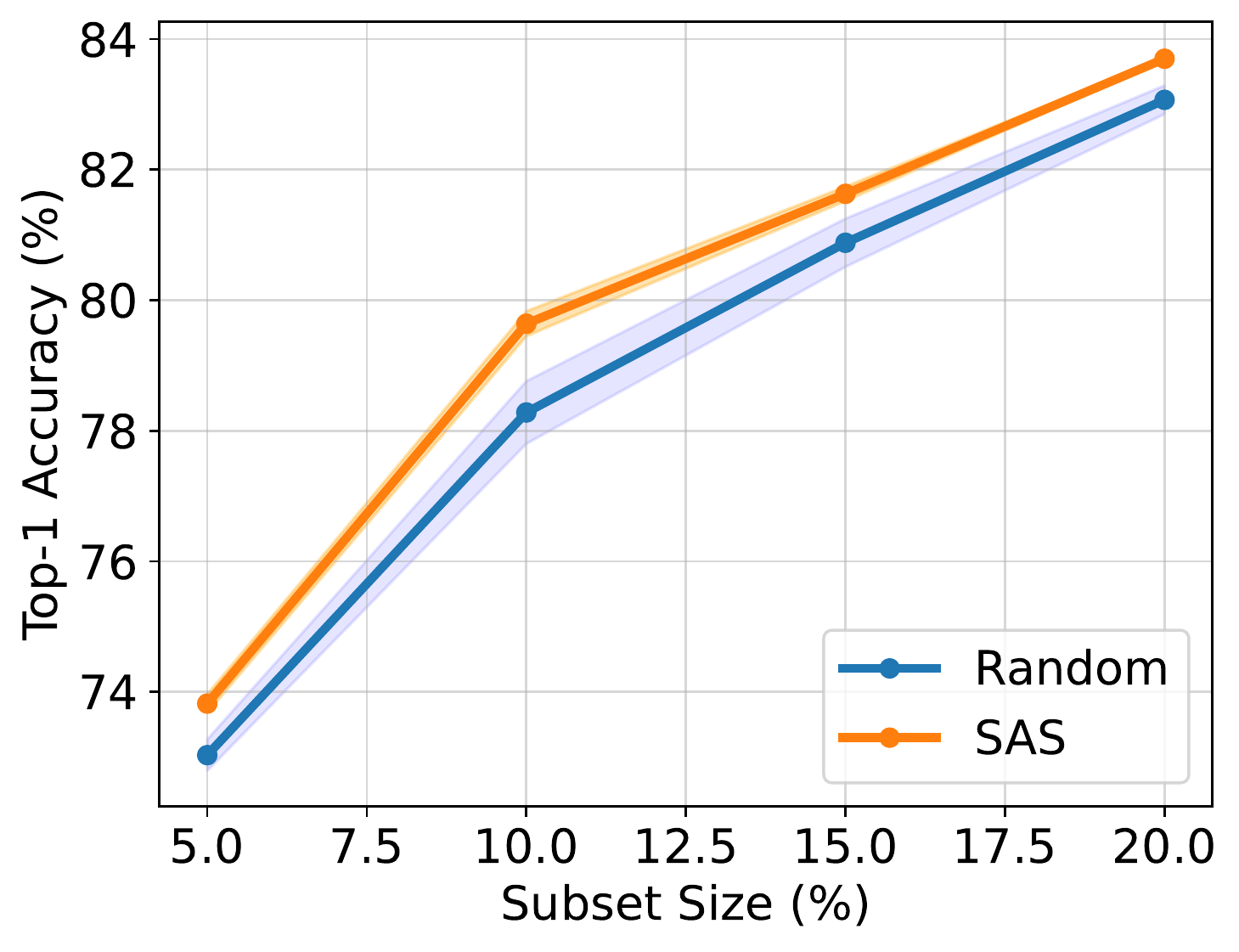} \label{fig:results_cifar10}} 
    \subfigure[CIFAR100]{\includegraphics[width=0.24\linewidth]{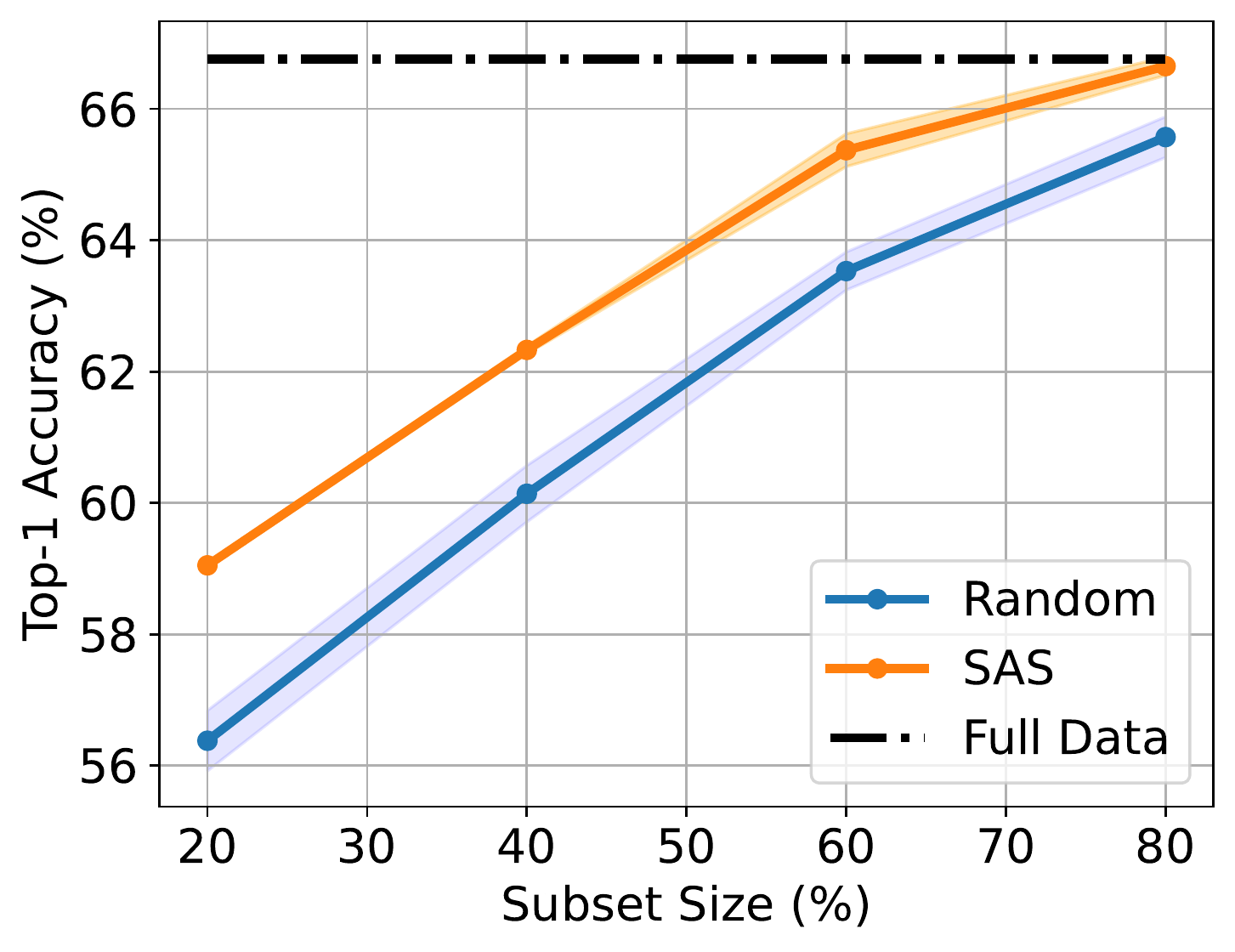} \label{fig:results_cifar100}} 
    \subfigure[STL10]{\includegraphics[width=0.24\linewidth]{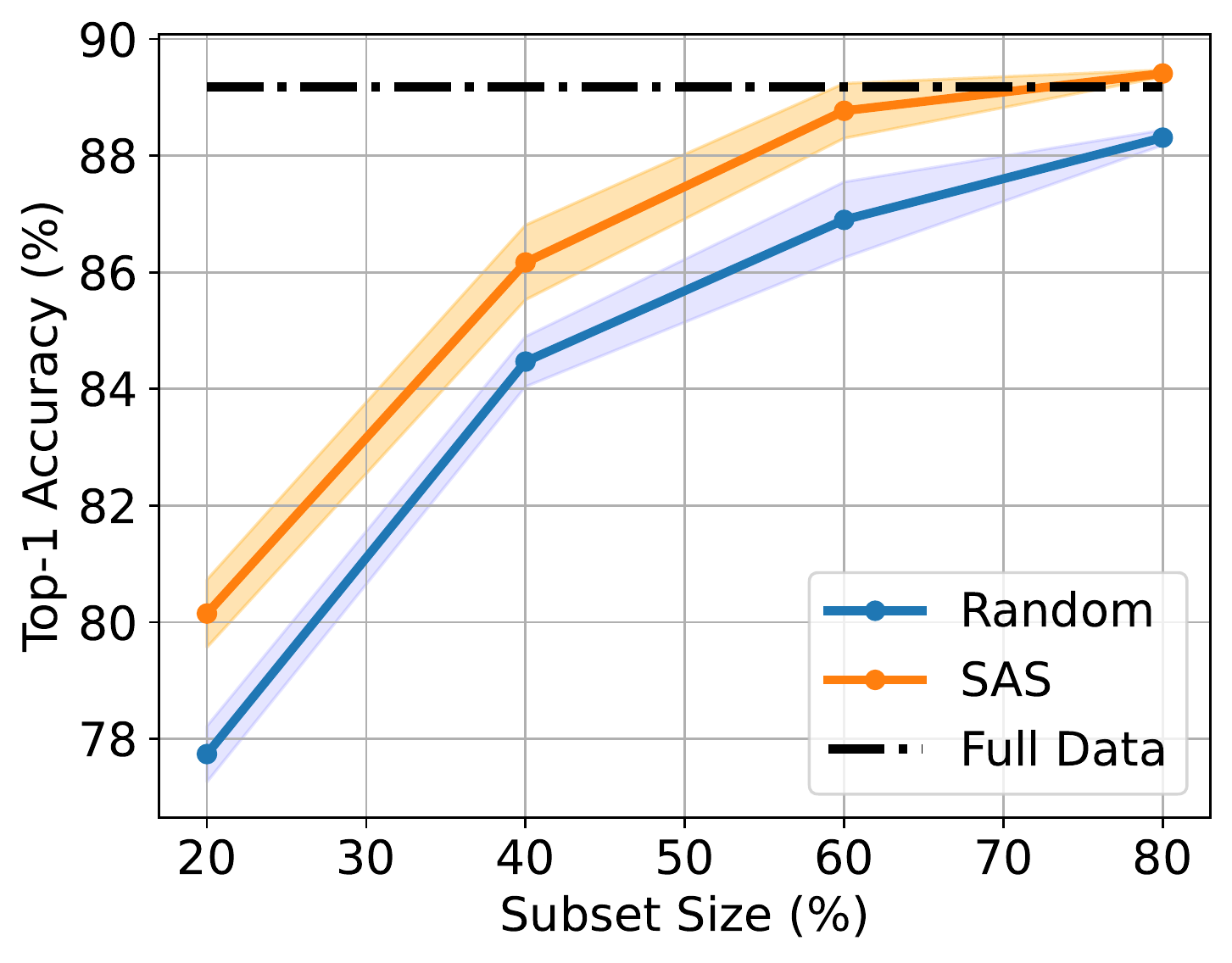} \label{fig:results_stl10}}
    \subfigure[TinyImageNet (ResNet-18)]{\includegraphics[width=0.24\linewidth]{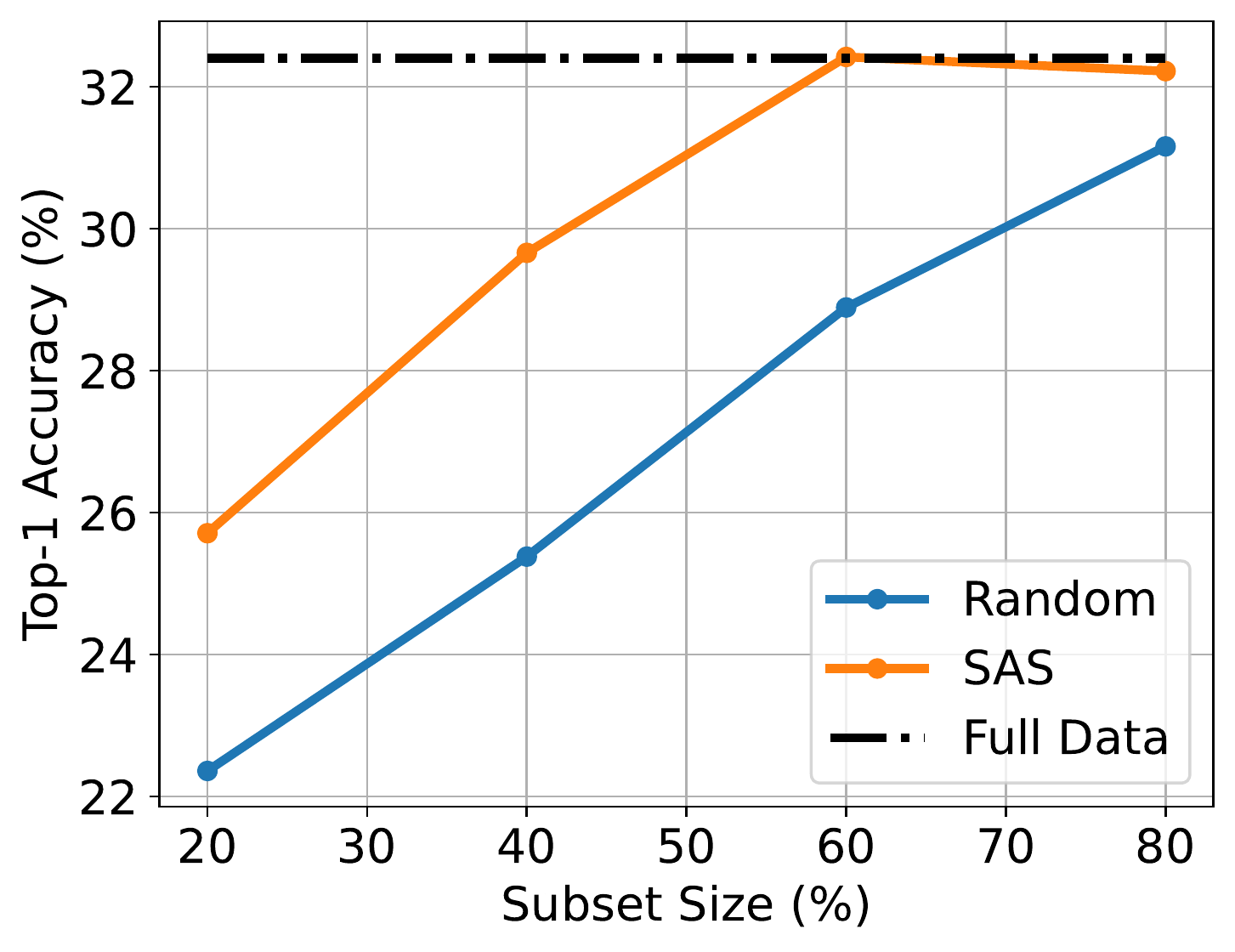} \label{fig:results_tinyimagenet}}
    \vspace{-4mm}
    \caption{Downstream Classification Accuracy of \method\ Subsets vs. Random Subsets (reporting mean and std over 3 runs).}
    \label{fig:results}
    \vspace{-3mm}
\end{figure*}

Finally, we 
present our method, \method, which finds subsets that minimize expected augmentation distance or equivalently maximize expected augmentation similarity, by
approximately finding the latent classes and estimating the expected augmentation distance, without having the labels.

\vspace{-1mm} \textbf{Approximately Finding the Latent Classes.}
Problem \eqref{eq:submodular} requires selecting a subset from every class separately. Without the labels, we need to approximately find the latent classes.
In practice, one can find latent classes by clustering the representations of a model trained with contrastive SSL. This approach requires no extra information
and thus can generalize to contrastive learning in all domains. However, if an extra small subset of labeled data and a proxy model is available, we can find latent classes much more accurately.
Specifically, if a small subset of labels are available, a proxy model can be used to approximately find the latent classes. In our experiments, we show that having as small as 1\% of the labels, the pretrained CLIP \cite{radford2021learning} image encoder can be used to find the latent classes more accurately. 
Crucially, even without having access to any downstream labels, the pretrained CLIP can be used to find the latent classes.
In our experiments, we show that using CLIP's image and text encoders, we can match image embeddings from STL10 to the closest text embeddings from ImageNet labels to obtain approximate latent classes for STL10. In practice, any \textit{fine-grained} relevant set of labels provide a superior performance. This is because linearly separable representations for the fine-grained task will ensure linearly separable representations for the coarser-grained task.
This is a practical way to use \method\ for vision tasks as well as other domains with pretrained foundational models. \vspace{-1mm}


\textbf{Estimating the Expected Augmentation Similarity.}
Expected augmentation distance captures similarity of examples in the input space. However, as pixel space is extremely high-dimensional, nearly all expected augmentation distances will be very large and extremely sensitive to small noise. Instead, using a proxy model can better capture the semantic similarities in practice.
Note that the proxy model does not necessarily have to be the same as the model being trained with SSL. Indeed, the proxy model can be much smaller than the model being trained or can be partially trained with similar augmentations, as we confirm experimentally.
Having a proxy model $f_p$, for all $\x_i, \x_j \in V_k$, we estimate \textit{expected augmentation similarity}, i.e., $C - d_{i, j}$ in Eq. \eqref{eq:submodular} by $s_{i,j} = \left< f_p(\x_i), f_p(\x_j)\right>$. The pseudocode of \method\ is illustrated in Alg. \ref{alg}. \looseness=-1 

\vspace{-4mm}
\section{Experiments}
\vspace{-2mm}

\begin{figure*}[t]
    \centering
    \subfigure[BYOL (STL10)]{\includegraphics[width=0.24\linewidth]{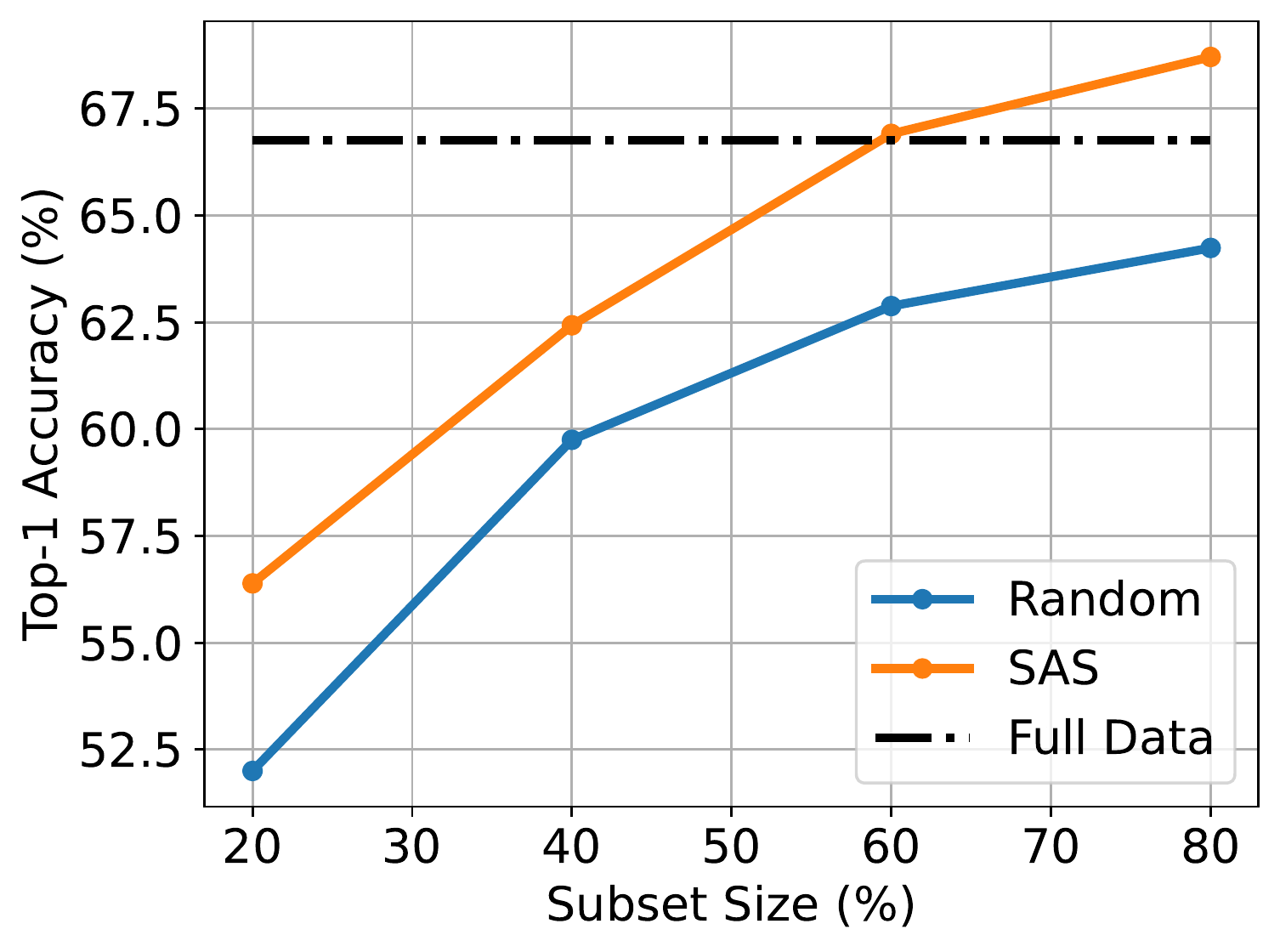} \label{fig:ablation_byol}}
    \subfigure[SimSiam (CIFAR100)]{\includegraphics[width=0.24\linewidth]{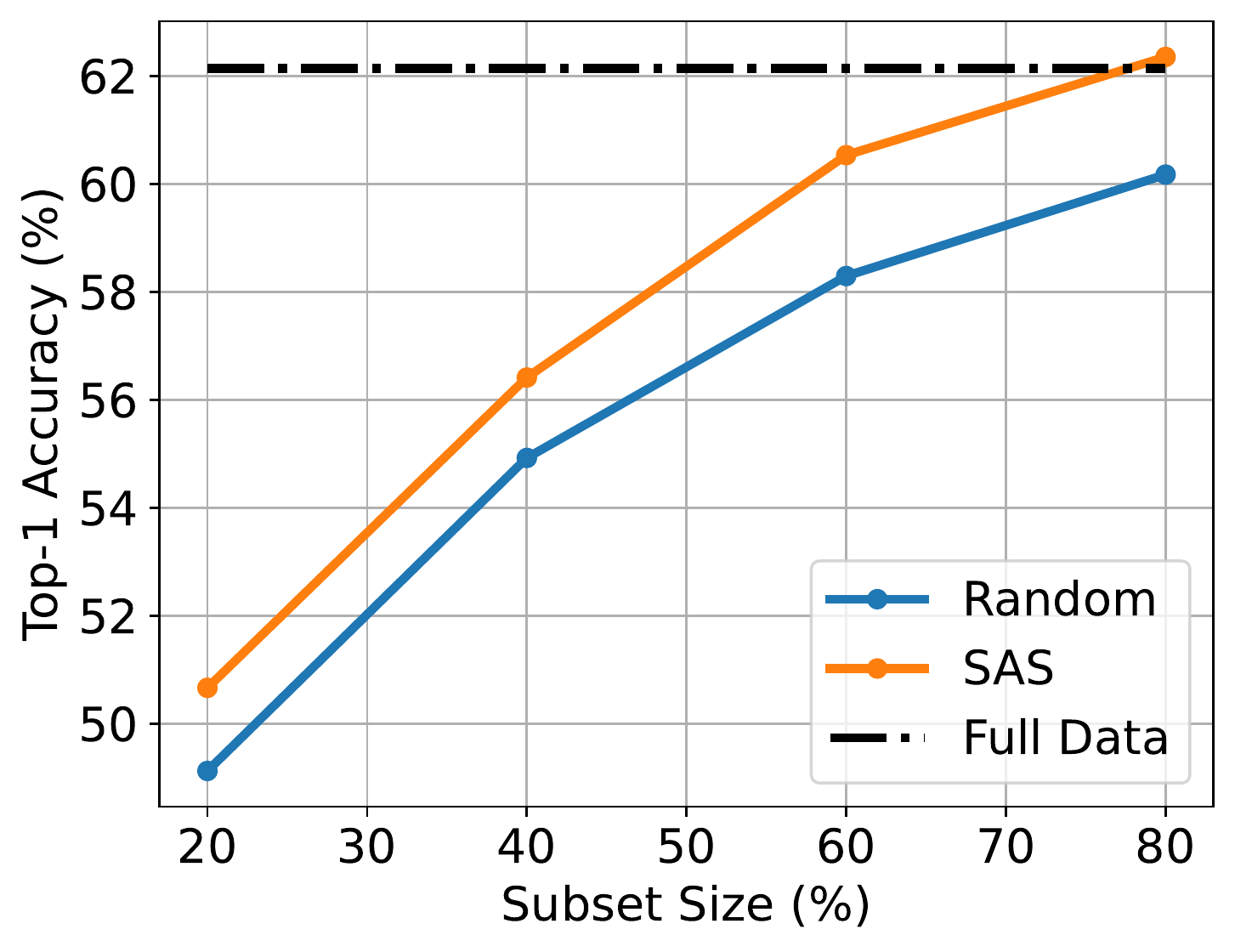} \label{fig:ablation_simsiam}}
    \subfigure[MoCo (CIFAR100)]{\includegraphics[width=0.24\linewidth]{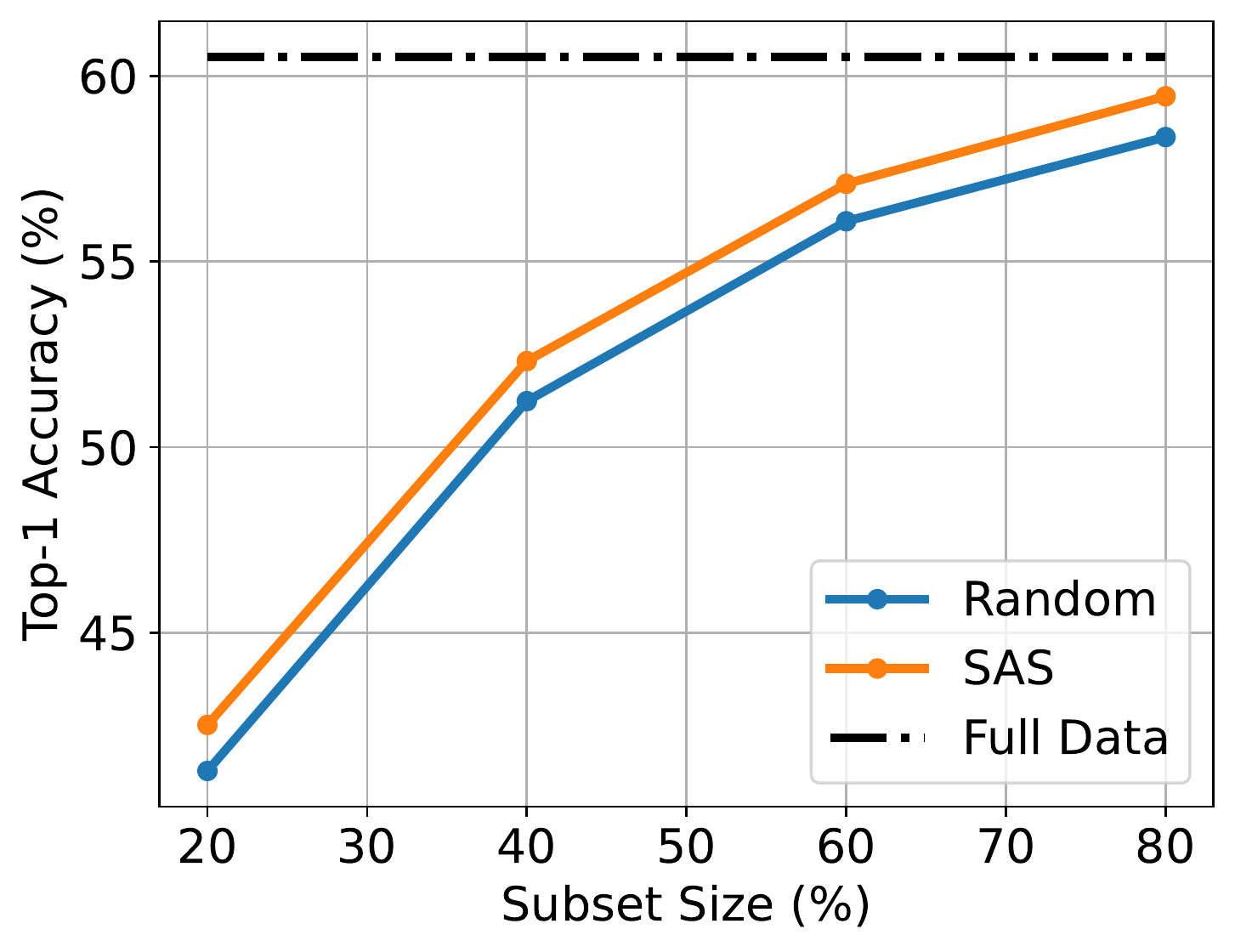} \label{fig:ablation_moco}}
    \vspace{-3mm}
    \caption{Evaluating \method\  on other contrastive learning methods (training a ResNet-18).}
    \label{fig:other_cl}
    \vspace{-3mm}
\end{figure*}

\begin{figure*}[t]
    \centering
    \subfigure[Effect of approximate latent classes (CIFAR100)]{\includegraphics[width=0.24\linewidth]{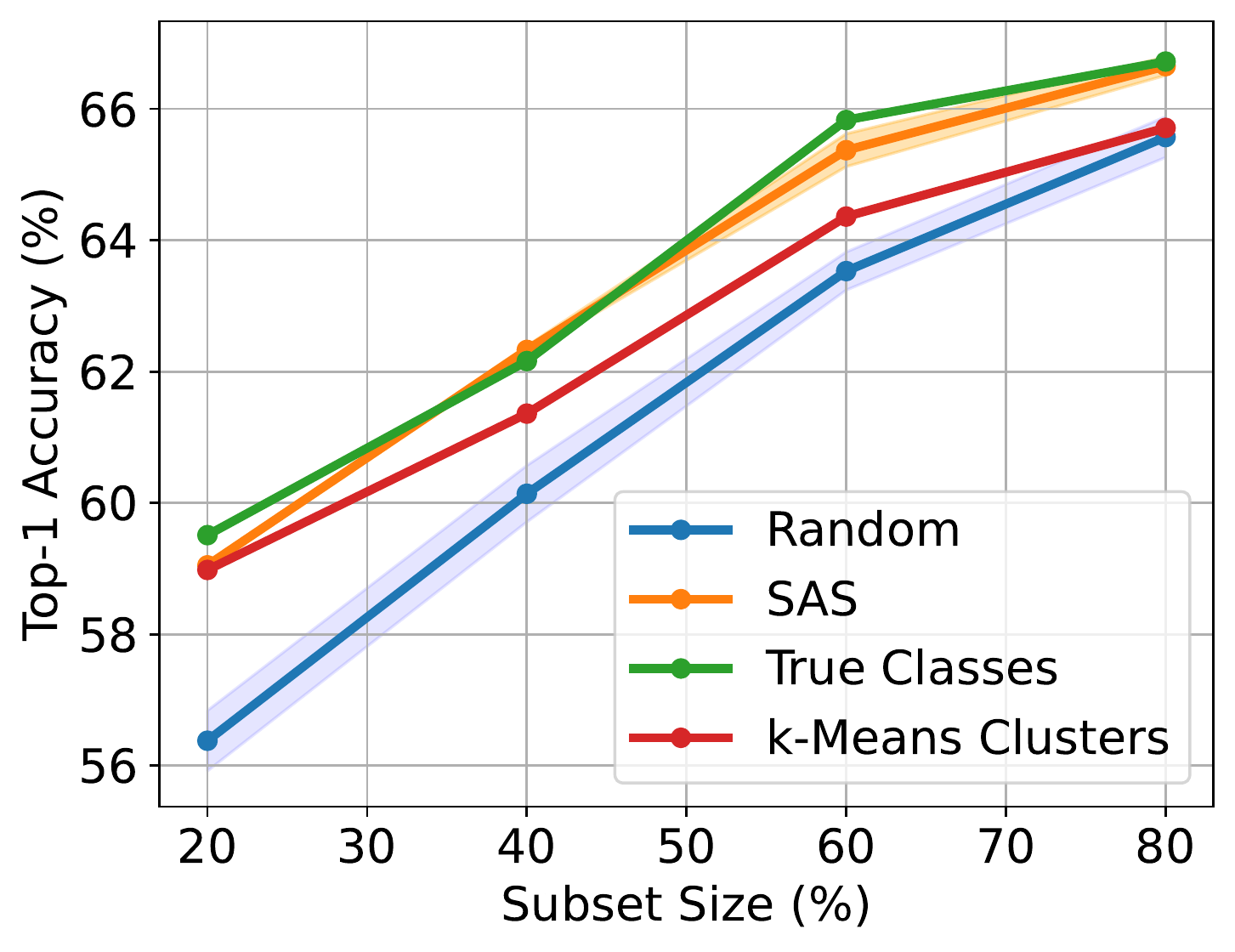} \label{fig:ablation_clustering}} 
    \subfigure[ImageNet labels (STL10)]{\includegraphics[width=0.24\linewidth]{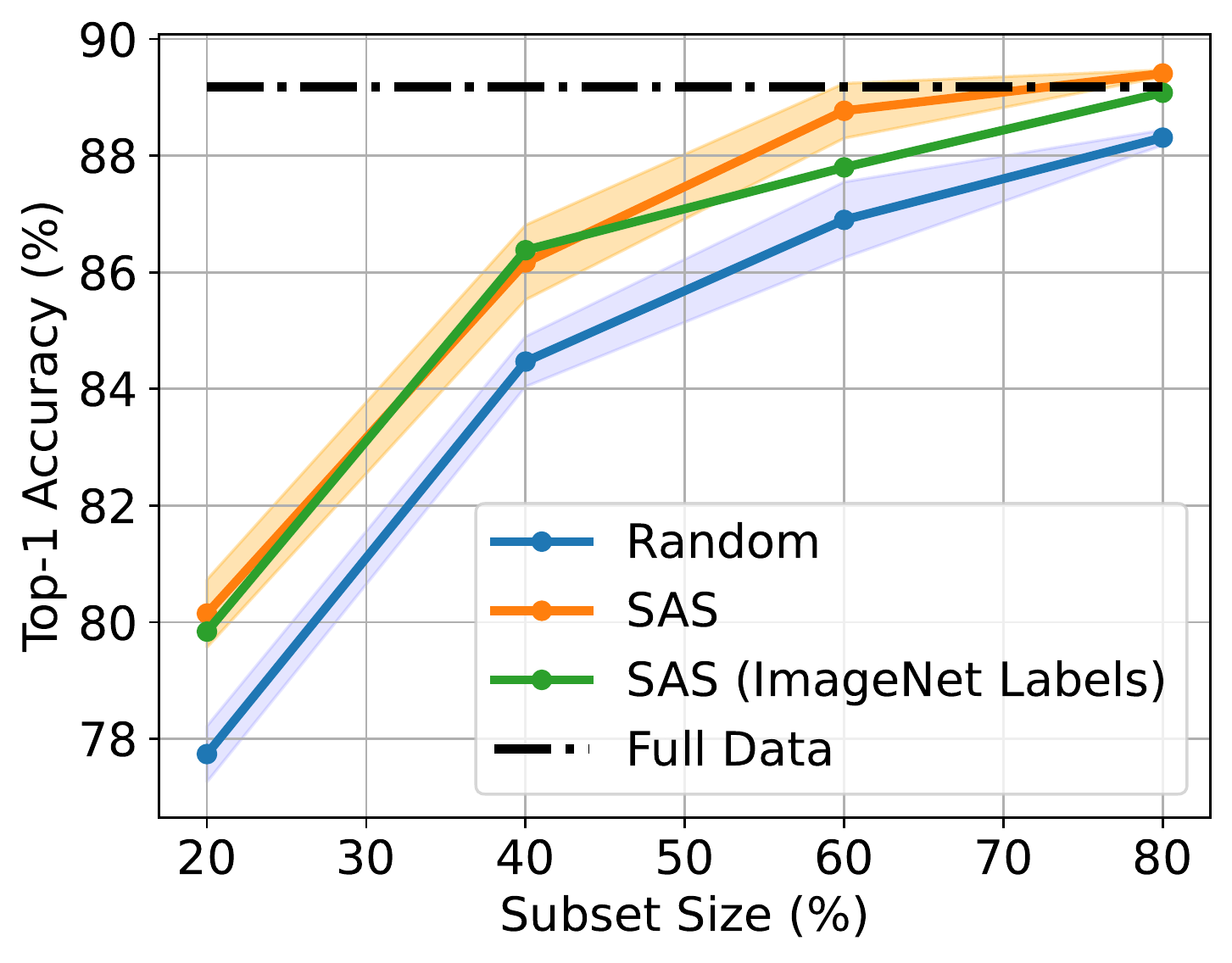} \label{fig:ablation_imgnet}}
    \subfigure[Effect of proxy model \newline
    (CIFAR100)]{\includegraphics[width=0.24\linewidth]{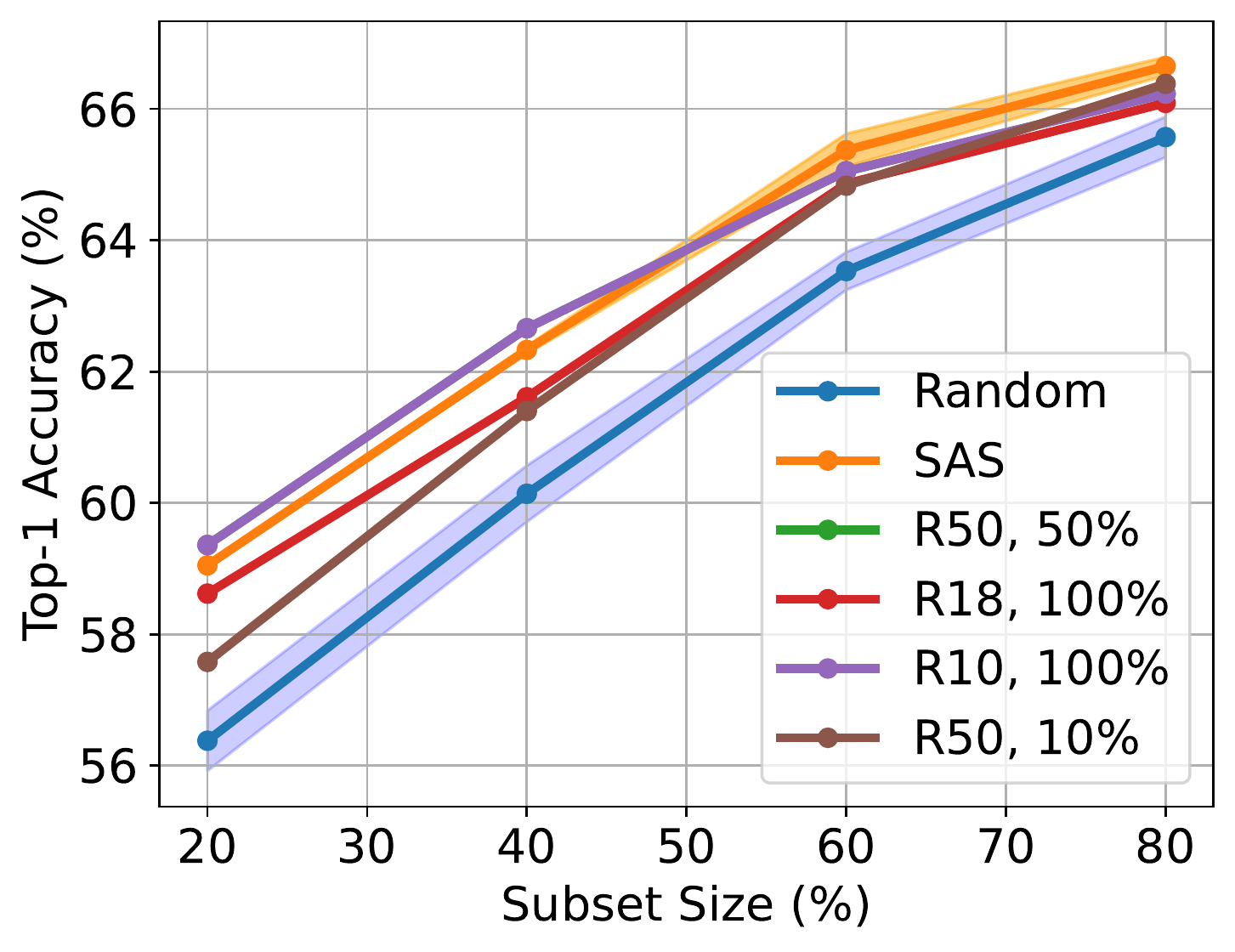} \label{fig:ablation_proxy}}
    \subfigure[\!Easy examples for SL are important for SSL \!(CIFAR100)]{\includegraphics[width=0.24\linewidth]{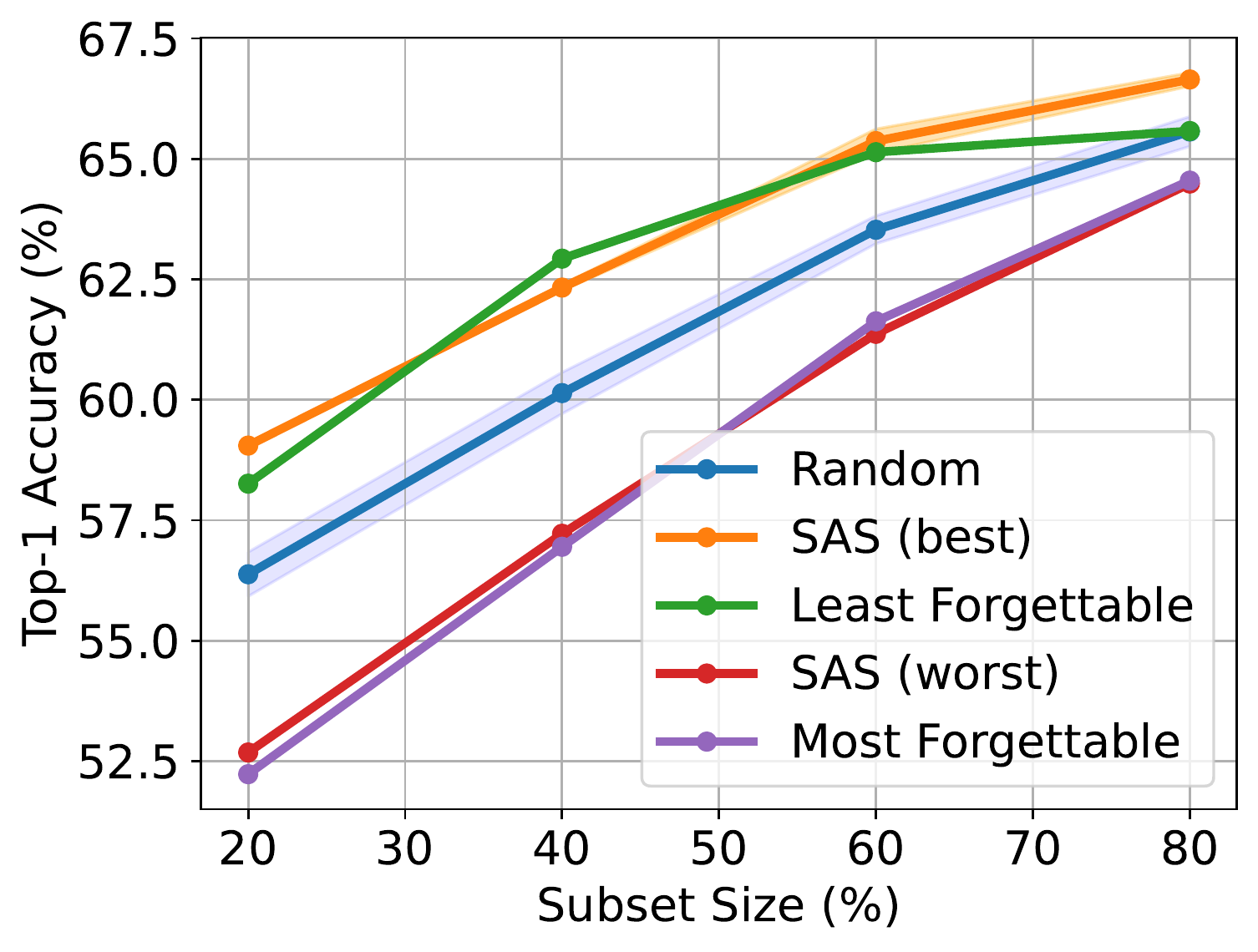} \label{fig:ablation_easy}} 
    \vspace{-4mm}
    \caption{Ablation study on CIFAR100 and STL10.}
    \label{fig:ablation}
    \vspace{-3mm}
\end{figure*}

In this section, we first evaluate the downstream generalization performance of the models trained by contrastive learning on the subsets found by \method~vs. random subsets,
on CIFAR10, CIFAR100 \cite{cifar}, STL10 \cite{stl10} and TinyImageNet \cite{imagenet}. Then, we do an extensive ablation study on the effect of the approximate latent classes, and the proxy model used to estimate expected augmentation distance. 
Finally, we investigate the relation of these subsets to subsets that are important for supervised learning.

 
\vspace{-1mm}\textbf{Training Setup}
We use SimCLR \cite{cl_simclr} as the contrastive learning method to train ResNet-50 \cite{resnet} as the encoder architecture and
a 2-layer MLP to project the representation to a 128-dimensional latent space. We use InfoNCE with temperature 
as our loss. Following the standard training pipeline in \cite{cl_chuang_debiased_2020, cl_robinson_hard_neg_2020} we train for 400 epochs using the Adam optimizer with a learning rate of $0.001$. Due to computational constraints, we use ResNet-18 as the encoder architecture for TinyImageNet and only have a single run per subset size. We also evaluate \method\ on other contrastive learning methods, namely BYOL \cite{grill2020bootstrap}, SimSiam \cite{cl_simsiam} and MoCo \cite{cl_he_momentum_2020}, using ResNet-18 as the encoder architecture.

\vspace{-1.5mm}\textbf{Data Augmentation} For data augmentations, we use 
random crop, random resizing, random horizontal flips and color distortion, as is done in \cite{cl_simclr}.

\vspace{-2mm} \textbf{Evaluation.} For evaluation, we use the widely used linear evaluation protocol \cite{cl_simclr, cl_chuang_debiased_2020}. That is, we train a linear classifier using the learned representations of the training examples and their labels. Then, we evaluate the performance of the linear classifier on the test set representations and their corresponding labels. 
To ensure fair comparison, we compare \method\ subsets with random subsets of the same size sampled from the same approximate latent classes.\looseness=-1

\vspace{-3mm}
\subsection{Downstream Generalization Performance}
\vspace{-1.5mm}

First, we evaluate the downstream generalization performance of the model pre-trained on subsets of different sizes found by \method~vs. random subsets of the same size. Here, we use a pre-trained ResNet-50 as the proxy to calculate $s_{ij}$, as discussed in Sec. \ref{sec:practice}.
For CIFAR100 and STL10, we consider all $s_{i, j} > 0$ and for CIFAR10 we consider all $s_{i, j} > 0.5$. As examples in CIFAR10 are generally more similar to each other, a larger threshold helps identifying representative examples better.
To approximately find the latent classes, we 
train a linear classifier on the CLIP representations of the training data with a small randomly selected subset of training labels. In particular, for CIFAR10 and CIFAR100, we use 1\% of the labels of training examples selected at random, and for STL10, we use all the labels ($<$ 5\%) available labels.
We use the trained linear classifiers to predict the latent class for all the training examples.
In our ablation studies, we evaluate the performance when finding latent classes in other ways. \looseness=-1



\textbf{SimCLR} 
Fig. \ref{fig:results} shows that 
training with SimCLR on subsets of various sizes found by \method\ allows outperform random subsets by over { 3\%} on CIFAR100 and STL10, and by up to {2\%} on CIFAR10. Critically, comparing the performance of the subsets with that of the full data, we can see that for CIFAR100, 20\% of examples and for STL10 and TinyImageNet, 40\% of examples, can be safely discarded without affecting downstream accuracy.

\vspace{-2mm}\textbf{Other Contrastive Learning Methods.} We validate that \method\ can effectively find examples that contribute the most to contrastive learning across  variety of contrastive learning methods. For these experiments, we train a ResNet-18 using BYOL \cite{cl_grill_bootstrap_2020}, MoCo \cite{cl_he_momentum_2020} and SimSiam \cite{cl_simsiam}. Fig. \ref{fig:ablation_byol} shows that training with BYOL on subsets of various sizes found by \method\ from STL10 outperforms random subsets by more than 3\%. Interestingly, with BYOL, subsets of size 80\% outperform training on the full data by {2\%}. We also show that \method\  allows us to discard 20\% of examples on CIFAR100 when using SimSiam (Fig. \ref{fig:ablation_simsiam}) and to achieve nearly 2\% improvement over random subsets when using MoCo. (Fig. \ref{fig:ablation_moco}). \looseness=-1

\vspace{-3mm}\subsection{Ablation Study}\vspace{-1.5mm}
Next, we conduct an extensive ablation study on the effect of the approximate latent classes,
and the proxy model used to estimate expected augmentation distance.

\textbf{Finding Approximate Latent Classes.} Fig. \ref{fig:ablation_clustering} compares the downstream performance on CIFAR100 when latent classes are obtained by training a linear classifier using 1\% labeled training data on CLIP representations, to that of using the ground-truth class labels, and $k$-means clustering on the representations of a pretrained model. We see that approximately finding the latent classes using 1\% of the labels works nearly as well as ground-truth labels. Notably, while the accuracy of the linear classifier trained with 1\% of the labels of CIFAR100 is only 70.8\%, this does not negatively affect the quality of subsets found by \method. The latent classes help us avoid confusing examples that are similar to examples across many latent classes; thus, even with relatively inaccurate latent classes, such examples can be filtered.
Moreover, in the absence of any labels, using $k$-means clustering on the on the representations of a pretrained model performs equally well for smaller subsets and still provides a significant improvement for larger  subsets. 

Next, we consider 
using a different set of labels than the original labels of the training data to find the latent classes. In particular, we use a pretrained CLIP to label STL10 images by ImageNet labels, using the zero-shot approach. That is, we match every image in STL10 to one of the ImageNet labels, by finding the CLIP text embedding of the ImageNet label that is most similar to the CLIP image embedding.
Fig. \ref{fig:ablation_imgnet} compares the downstream performance on STL10, when using ImageNet labels to find latent classes using a zero-shot approach to that of using the available ($<5\%$) STL10 labels to train a linear classifier on CLIP image representations.
Notably, no label information about STL is used in the first case.
The results clearly shows how \method\  can entirely avoid the use of labels for approximating the latent classes. Crucially, any relevant and potentially finer-grained set of labels are enough to approximately find the latent classes and achieve a superior downstream performance. 

\vspace{-1mm}
\textbf{Using Proxy Models to Estimate Expected Augmentation Distance.} Fig. \ref{fig:ablation_proxy} shows estimating augmentation distance using various proxy models, such as a ResNet-50 that is partially trained for as few as 10\% of epochs as well as smaller models such as a pre-trained ResNet-10, achieves a very similar downstream performance to that of using a fully pre-trained ResNet-50.

\vspace{-3mm}
\subsection{Investigating subsets found by \method}\vspace{-2mm}

\begin{figure}
\centering
\vspace{-5mm}
    \subfigure[t-SNE 
    of \textit{bed} (pairs with small $d_{\x\x'}\!$ are connected)]{
      \includegraphics[width=0.45\linewidth, keepaspectratio]{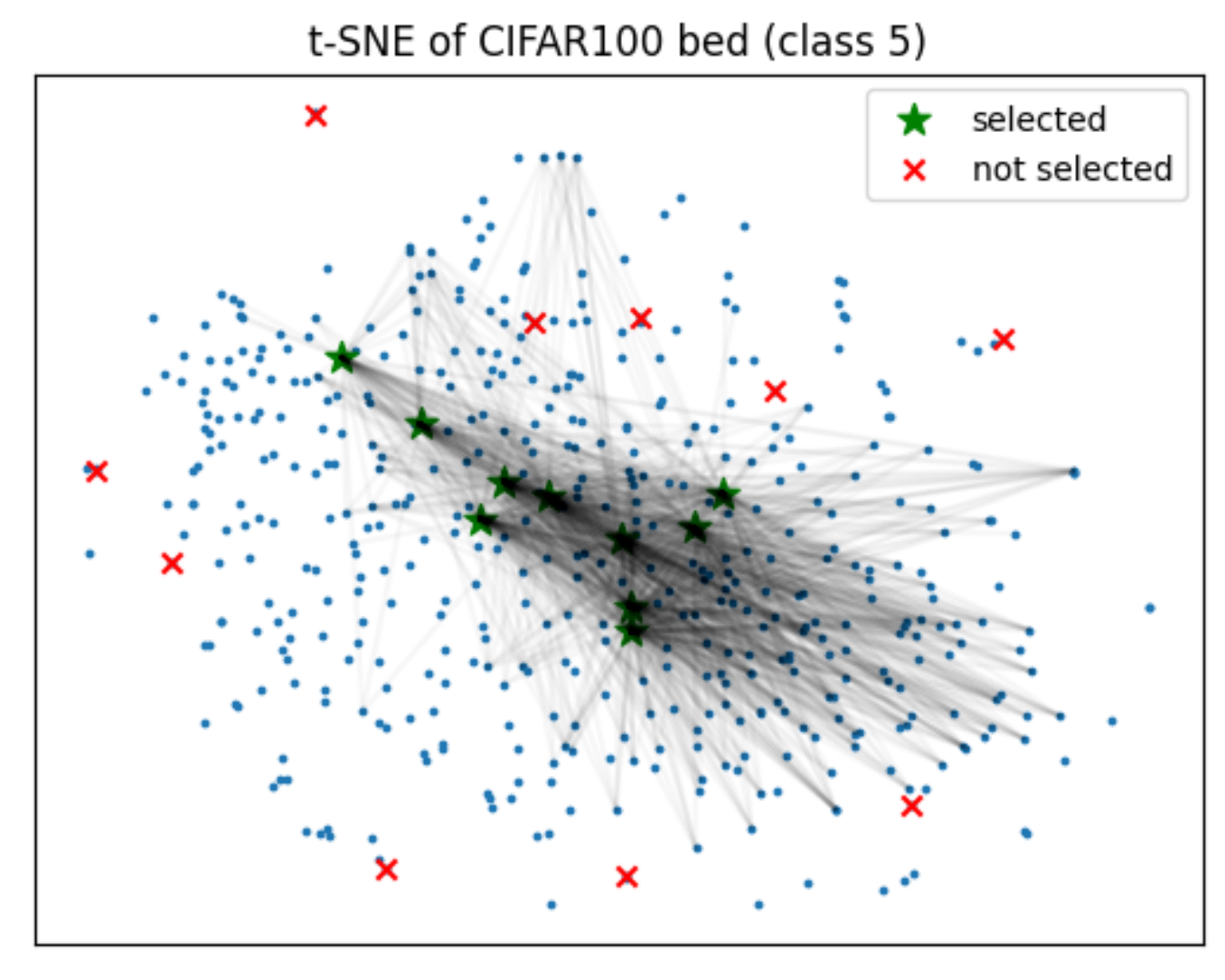} \label{fig:cifar100_viz_tsne}}
    \subfigure[Examples from \textit{bicycle}]{
  \includegraphics[width=0.45\linewidth, keepaspectratio]{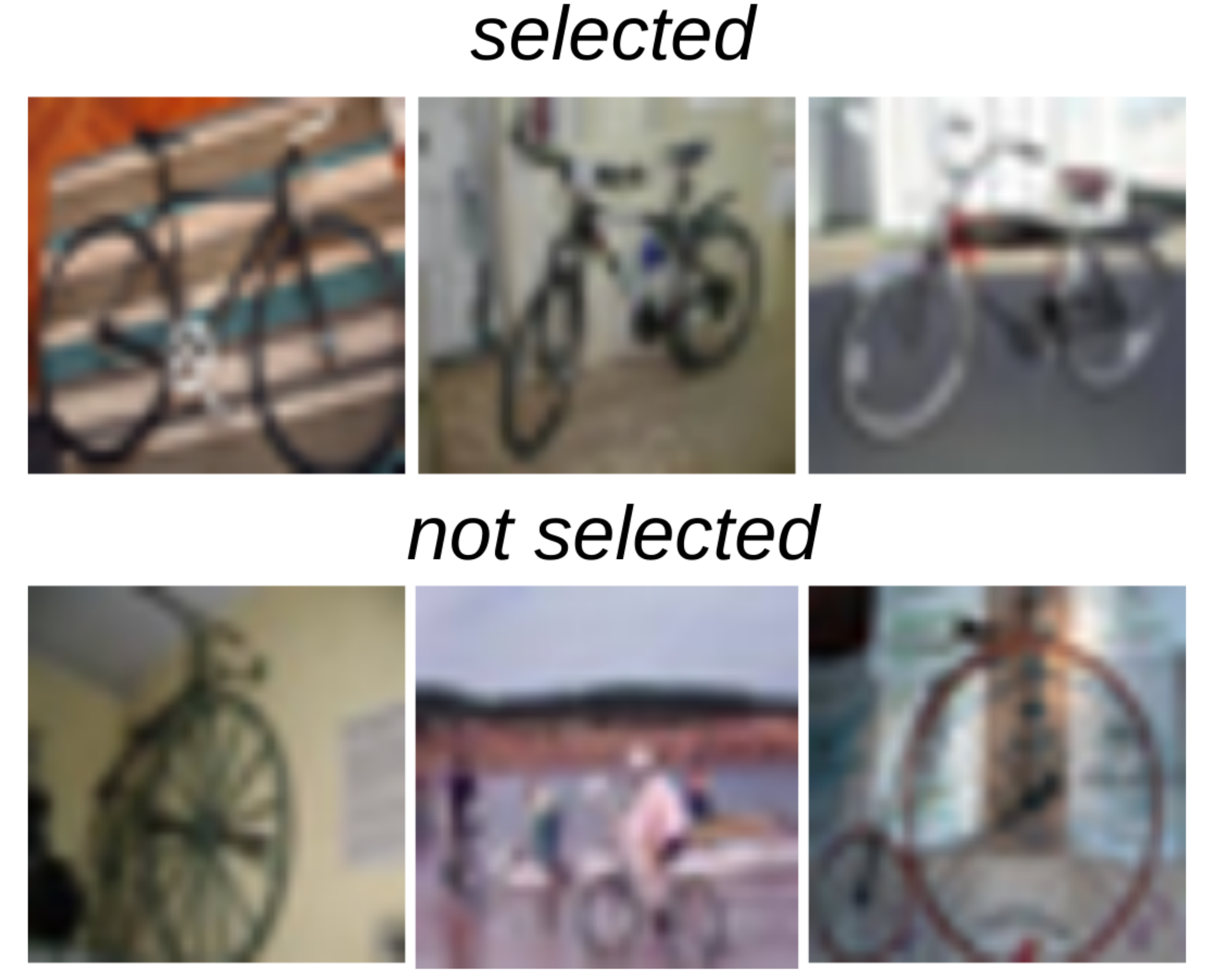} \label{fig:cifar100_viz_examples}}
      \vspace{-4mm}
    \caption{Visualizing selected examples from CIFAR100}
    \label{fig:cifar100_viz}
    \vspace{-8mm}
\end{figure}

\textbf{Visualization.} Fig. \ref{fig:cifar100_viz_tsne} use t-SNE to visualize examples that are selected by \method\ vs those that are not selected, from the class ``bed" in CIFAR100. Examples with small expected augmentation distance to selected and not selected examples are connected. We see that the selected examples have small 
distance to many other examples in the class. 
Fig. \ref{fig:cifar100_viz_examples}, illustrates some examples that are selected and not selected from the ``bicycle" class.
We see that the selected examples are representatives of the whole class, while those not selected present uncommon poses or views of the object. \looseness=-1

\textbf{Easy Examples are the Most Important.} Finally, we use the forgetting score \cite{dp_toneva_forgettability_2019}, i.e. the number of times an examples is misclassified after being correctly classified during \textit{supervised} learning, to quantify the difficulty of an example. 
Importantly, least forgettable examples that can be safely discarded from supervised learning without harming the accuracy \cite{dp_toneva_forgettability_2019}.
Fig. \ref{fig:ablation_easy} shows that least forgettable examples can considerably outperform the random baseline and achieve a comparable performance to \method\ for smaller subsets. On the other hand, the most forgettable examples that are most beneficial for supervised learning, perform significantly worse than the random baseline and similar to the subsets deemed worst by \method. This illustrates how the subsets that contribute the most to contrastive learning are the least beneficial for supervised learning and vice-a-versa. Fig. \ref{fig:fscore_confidence} in Appendix \ref{appendix:experiments} further shows that subsets found by \method\ have low forgetting score and high confidence, in expectation. That is, they are easy for supervised learning.
Effectively, the most important subsets for SSL are least important for supervised learning.

\vspace{-4mm}\section{Conclusion}\vspace{-2mm}
We identified subsets of examples that contribute the most to contrastive SSL. Theoretically, we characterized important subsets for contrastive learning with rigorous generalization guarantees for downstream performance.  Empirically, we showed that using our method 20\% - 40\% examples can be discarded on CIFAR100, STL10 and TinyImageNet, observing no loss and even improvement in downstream accuracy. Surprisingly, we discovered that these important subsets are the least informative for supervised learning.

\vspace{-2mm}
\textbf{Acknowledgment.} This research was supported by 
the National Science Foundation CAREER
Award 2146492.\looseness=-1



\bibliography{references}

\begin{thebibliography}{37}
\providecommand{\natexlab}[1]{#1}
\providecommand{\url}[1]{\texttt{#1}}
\expandafter\ifx\csname urlstyle\endcsname\relax
  \providecommand{\doi}[1]{doi: #1}\else
  \providecommand{\doi}{doi: \begingroup \urlstyle{rm}\Url}\fi

\bibitem[Arora et~al.(2019)Arora, Khandeparkar, Khodak, Plevrakis, and
  Saunshi]{arora2019theoretical}
Arora, S., Khandeparkar, H., Khodak, M., Plevrakis, O., and Saunshi, N.
\newblock A theoretical analysis of contrastive unsupervised representation
  learning.
\newblock \emph{arXiv preprint arXiv:1902.09229}, 2019.

\bibitem[Buchbinder et~al.(2015)Buchbinder, Feldman, Seffi, and
  Schwartz]{buchbinder2015tight}
Buchbinder, N., Feldman, M., Seffi, J., and Schwartz, R.
\newblock A tight linear time (1/2)-approximation for unconstrained submodular
  maximization.
\newblock \emph{SIAM Journal on Computing}, 44\penalty0 (5):\penalty0
  1384--1402, 2015.

\bibitem[Chen et~al.(2020)Chen, Kornblith, Norouzi, and Hinton]{cl_simclr}
Chen, T., Kornblith, S., Norouzi, M., and Hinton, G.
\newblock A {Simple} {Framework} for {Contrastive} {Learning} of {Visual}
  {Representations}.
\newblock February 2020.
\newblock \doi{10.48550/arXiv.2002.05709}.
\newblock URL \url{https://arxiv.org/abs/2002.05709v3}.

\bibitem[Chen \& He(2020)Chen and He]{cl_simsiam}
Chen, X. and He, K.
\newblock Exploring simple siamese representation learning, 2020.

\bibitem[Chen \& He(2021)Chen and He]{chen2021exploring}
Chen, X. and He, K.
\newblock Exploring simple siamese representation learning.
\newblock In \emph{Proceedings of the IEEE/CVF Conference on Computer Vision
  and Pattern Recognition}, pp.\  15750--15758, 2021.

\bibitem[Chuang et~al.(2020)Chuang, Robinson, Lin, Torralba, and
  Jegelka]{cl_chuang_debiased_2020}
Chuang, C.-Y., Robinson, J., Lin, Y.-C., Torralba, A., and Jegelka, S.
\newblock Debiased {Contrastive} {Learning}.
\newblock In \emph{Advances in {Neural} {Information} {Processing} {Systems}},
  volume~33, pp.\  8765--8775. Curran Associates, Inc., 2020.
\newblock URL
  \url{https://proceedings.neurips.cc/paper/2020/hash/63c3ddcc7b23daa1e42dc41f9a44a873-Abstract.html}.

\bibitem[Coates et~al.(2011{\natexlab{a}})Coates, Ng, and
  Lee]{coates2011analysis}
Coates, A., Ng, A., and Lee, H.
\newblock An analysis of single-layer networks in unsupervised feature
  learning.
\newblock In \emph{Proceedings of the fourteenth international conference on
  artificial intelligence and statistics}, pp.\  215--223. JMLR Workshop and
  Conference Proceedings, 2011{\natexlab{a}}.

\bibitem[Coates et~al.(2011{\natexlab{b}})Coates, Ng, and Lee]{stl10}
Coates, A., Ng, A., and Lee, H.
\newblock An analysis of single-layer networks in unsupervised feature
  learning.
\newblock In Gordon, G., Dunson, D., and Dudík, M. (eds.), \emph{Proceedings
  of the Fourteenth International Conference on Artificial Intelligence and
  Statistics}, volume~15 of \emph{Proceedings of Machine Learning Research},
  pp.\  215--223, Fort Lauderdale, FL, USA, 11--13 Apr 2011{\natexlab{b}}.
  PMLR.
\newblock URL \url{https://proceedings.mlr.press/v15/coates11a.html}.

\bibitem[Coleman et~al.(2020)Coleman, Yeh, Mussmann, Mirzasoleiman, Bailis,
  Liang, Leskovec, and Zaharia]{coleman2020selection}
Coleman, C., Yeh, C., Mussmann, S., Mirzasoleiman, B., Bailis, P., Liang, P.,
  Leskovec, J., and Zaharia, M.
\newblock Selection via proxy: Efficient data selection for deep learning.
\newblock In \emph{International Conference on Learning Representations
  (ICLR)}, 2020.

\bibitem[Deng et~al.(2009)Deng, Dong, Socher, Li, Li, and Fei-Fei]{imagenet}
Deng, J., Dong, W., Socher, R., Li, L.-J., Li, K., and Fei-Fei, L.
\newblock Imagenet: A large-scale hierarchical image database.
\newblock In \emph{2009 IEEE Conference on Computer Vision and Pattern
  Recognition}, pp.\  248--255, 2009.
\newblock \doi{10.1109/CVPR.2009.5206848}.

\bibitem[Grill et~al.(2020{\natexlab{a}})Grill, Strub, Altch{\'e}, Tallec,
  Richemond, Buchatskaya, Doersch, Avila~Pires, Guo, Gheshlaghi~Azar,
  et~al.]{grill2020bootstrap}
Grill, J.-B., Strub, F., Altch{\'e}, F., Tallec, C., Richemond, P.,
  Buchatskaya, E., Doersch, C., Avila~Pires, B., Guo, Z., Gheshlaghi~Azar, M.,
  et~al.
\newblock Bootstrap your own latent-a new approach to self-supervised learning.
\newblock \emph{Advances in neural information processing systems},
  33:\penalty0 21271--21284, 2020{\natexlab{a}}.

\bibitem[Grill et~al.(2020{\natexlab{b}})Grill, Strub, Altché, Tallec,
  Richemond, Buchatskaya, Doersch, Pires, Guo, Azar, Piot, Kavukcuoglu, Munos,
  and Valko]{cl_grill_bootstrap_2020}
Grill, J.-B., Strub, F., Altché, F., Tallec, C., Richemond, P.~H.,
  Buchatskaya, E., Doersch, C., Pires, B.~A., Guo, Z.~D., Azar, M.~G., Piot,
  B., Kavukcuoglu, K., Munos, R., and Valko, M.
\newblock Bootstrap your own latent: {A} new approach to self-supervised
  {Learning}, September 2020{\natexlab{b}}.
\newblock URL \url{http://arxiv.org/abs/2006.07733}.
\newblock arXiv:2006.07733 [cs, stat].

\bibitem[HaoChen et~al.(2021)HaoChen, Wei, Gaidon, and
  Ma]{cl_provable_guarantees}
HaoChen, J.~Z., Wei, C., Gaidon, A., and Ma, T.
\newblock Provable guarantees for self-supervised deep learning with spectral
  contrastive loss, 2021.
\newblock URL \url{https://arxiv.org/abs/2106.04156}.

\bibitem[He et~al.(2016)He, Zhang, Ren, and Sun]{resnet}
He, K., Zhang, X., Ren, S., and Sun, J.
\newblock Deep residual learning for image recognition.
\newblock In \emph{2016 IEEE Conference on Computer Vision and Pattern
  Recognition (CVPR)}, pp.\  770--778, 2016.
\newblock \doi{10.1109/CVPR.2016.90}.

\bibitem[He et~al.(2020)He, Fan, Wu, Xie, and Girshick]{cl_he_momentum_2020}
He, K., Fan, H., Wu, Y., Xie, S., and Girshick, R.
\newblock Momentum {Contrast} for {Unsupervised} {Visual} {Representation}
  {Learning}, March 2020.
\newblock URL \url{http://arxiv.org/abs/1911.05722}.
\newblock arXiv:1911.05722 [cs].

\bibitem[Huang et~al.(2021)Huang, Yi, and Zhao]{huang2021towards}
Huang, W., Yi, M., and Zhao, X.
\newblock Towards the generalization of contrastive self-supervised learning.
\newblock \emph{arXiv preprint arXiv:2111.00743}, 2021.

\bibitem[Katharopoulos \& Fleuret(2018)Katharopoulos and
  Fleuret]{katharopoulos2018not}
Katharopoulos, A. and Fleuret, F.
\newblock Not all samples are created equal: Deep learning with importance
  sampling.
\newblock In \emph{International conference on machine learning}, pp.\
  2525--2534. PMLR, 2018.

\bibitem[Killamsetty et~al.(2021)Killamsetty, Durga, Ramakrishnan, De, and
  Iyer]{killamsetty2021grad}
Killamsetty, K., Durga, S., Ramakrishnan, G., De, A., and Iyer, R.
\newblock Grad-match: Gradient matching based data subset selection for
  efficient deep model training.
\newblock In \emph{International Conference on Machine Learning}, pp.\
  5464--5474. PMLR, 2021.

\bibitem[Krizhevsky \& Hinton(2009)Krizhevsky and Hinton]{cifar}
Krizhevsky, A. and Hinton, G.
\newblock Learning multiple layers of features from tiny images.
\newblock Technical Report~0, University of Toronto, Toronto, Ontario, 2009.

\bibitem[Krizhevsky et~al.(2009)Krizhevsky, Hinton,
  et~al.]{krizhevsky2009learning}
Krizhevsky, A., Hinton, G., et~al.
\newblock Learning multiple layers of features from tiny images.
\newblock 2009.

\bibitem[Le \& Yang(2015)Le and Yang]{tiny_imagenet}
Le, Y. and Yang, X.~S.
\newblock Tiny imagenet visual recognition challenge.
\newblock 2015.

\bibitem[Mindermann et~al.(2022)Mindermann, Brauner, Razzak, Sharma, Kirsch,
  Xu, Höltgen, Gomez, Morisot, Farquhar, and
  Gal]{dp_mindermann_prioritized_2022}
Mindermann, S., Brauner, J.~M., Razzak, M.~T., Sharma, M., Kirsch, A., Xu, W.,
  Höltgen, B., Gomez, A.~N., Morisot, A., Farquhar, S., and Gal, Y.
\newblock Prioritized {Training} on {Points} that are {Learnable}, {Worth}
  {Learning}, and not yet {Learnt}.
\newblock In \emph{Proceedings of the 39th {International} {Conference} on
  {Machine} {Learning}}, pp.\  15630--15649. PMLR, June 2022.
\newblock URL \url{https://proceedings.mlr.press/v162/mindermann22a.html}.
\newblock ISSN: 2640-3498.

\bibitem[Minoux(2005)]{minoux2005accelerated}
Minoux, M.
\newblock Accelerated greedy algorithms for maximizing submodular set
  functions.
\newblock In \emph{Optimization Techniques: Proceedings of the 8th IFIP
  Conference on Optimization Techniques W{\"u}rzburg, September 5--9, 1977},
  pp.\  234--243. Springer, 2005.

\bibitem[Mirzasoleiman et~al.(2016)Mirzasoleiman, Badanidiyuru, and
  Karbasi]{mirzasoleiman2016fast}
Mirzasoleiman, B., Badanidiyuru, A., and Karbasi, A.
\newblock Fast constrained submodular maximization: Personalized data
  summarization.
\newblock In \emph{International Conference on Machine Learning}, pp.\
  1358--1367. PMLR, 2016.

\bibitem[Mirzasoleiman et~al.(2020)Mirzasoleiman, Bilmes, and
  Leskovec]{dp-craig-mirzasoleiman20a}
Mirzasoleiman, B., Bilmes, J., and Leskovec, J.
\newblock Coresets for data-efficient training of machine learning models.
\newblock In III, H.~D. and Singh, A. (eds.), \emph{Proceedings of the 37th
  International Conference on Machine Learning}, volume 119 of
  \emph{Proceedings of Machine Learning Research}, pp.\  6950--6960. PMLR,
  13--18 Jul 2020.
\newblock URL \url{https://proceedings.mlr.press/v119/mirzasoleiman20a.html}.

\bibitem[Oord et~al.(2018)Oord, Li, and Vinyals]{oord2018representation}
Oord, A. v.~d., Li, Y., and Vinyals, O.
\newblock Representation learning with contrastive predictive coding.
\newblock \emph{arXiv preprint arXiv:1807.03748}, 2018.

\bibitem[Paul et~al.(2021)Paul, Ganguli, and Dziugaite]{dp_el2n_paul_deep_2021}
Paul, M., Ganguli, S., and Dziugaite, G.~K.
\newblock Deep {Learning} on a {Data} {Diet}: {Finding} {Important} {Examples}
  {Early} in {Training}.
\newblock In \emph{Advances in {Neural} {Information} {Processing} {Systems}},
  volume~34, pp.\  20596--20607. Curran Associates, Inc., 2021.
\newblock URL
  \url{https://proceedings.neurips.cc/paper/2021/hash/ac56f8fe9eea3e4a365f29f0f1957c55-Abstract.html}.

\bibitem[Pooladzandi et~al.(2022)Pooladzandi, Davini, and
  Mirzasoleiman]{pooladzandi2022adaptive}
Pooladzandi, O., Davini, D., and Mirzasoleiman, B.
\newblock Adaptive second order coresets for data-efficient machine learning.
\newblock In \emph{International Conference on Machine Learning}, pp.\
  17848--17869. PMLR, 2022.

\bibitem[Radford et~al.(2021)Radford, Kim, Hallacy, Ramesh, Goh, Agarwal,
  Sastry, Askell, Mishkin, Clark, et~al.]{radford2021learning}
Radford, A., Kim, J.~W., Hallacy, C., Ramesh, A., Goh, G., Agarwal, S., Sastry,
  G., Askell, A., Mishkin, P., Clark, J., et~al.
\newblock Learning transferable visual models from natural language
  supervision.
\newblock In \emph{International Conference on Machine Learning}, pp.\
  8748--8763. PMLR, 2021.

\bibitem[Robinson et~al.(2020)Robinson, Chuang, Sra, and
  Jegelka]{cl_robinson_hard_neg_2020}
Robinson, J., Chuang, C.-Y., Sra, S., and Jegelka, S.
\newblock Contrastive {Learning} with {Hard} {Negative} {Samples}.
\newblock October 2020.
\newblock \doi{10.48550/arXiv.2010.04592}.
\newblock URL \url{https://arxiv.org/abs/2010.04592v2}.

\bibitem[Saunshi et~al.(2019)Saunshi, Plevrakis, Arora, Khodak, and
  Khandeparkar]{cl_theory_saunshi_2019}
Saunshi, N., Plevrakis, O., Arora, S., Khodak, M., and Khandeparkar, H.
\newblock A theoretical analysis of contrastive unsupervised representation
  learning.
\newblock In Chaudhuri, K. and Salakhutdinov, R. (eds.), \emph{Proceedings of
  the 36th International Conference on Machine Learning}, volume~97 of
  \emph{Proceedings of Machine Learning Research}, pp.\  5628--5637. PMLR,
  09--15 Jun 2019.
\newblock URL \url{https://proceedings.mlr.press/v97/saunshi19a.html}.

\bibitem[Sorscher et~al.(2022)Sorscher, Geirhos, Shekhar, Ganguli, and
  Morcos]{dp_sorscher_beyond_2022}
Sorscher, B., Geirhos, R., Shekhar, S., Ganguli, S., and Morcos, A.~S.
\newblock Beyond neural scaling laws: beating power law scaling via data
  pruning, August 2022.
\newblock URL \url{http://arxiv.org/abs/2206.14486}.
\newblock arXiv:2206.14486 [cs, stat].

\bibitem[Swayamdipta et~al.(2020)Swayamdipta, Schwartz, Lourie, Wang,
  Hajishirzi, Smith, and Choi]{dp_swayamdipta_dataset_2020}
Swayamdipta, S., Schwartz, R., Lourie, N., Wang, Y., Hajishirzi, H., Smith,
  N.~A., and Choi, Y.
\newblock Dataset {Cartography}: {Mapping} and {Diagnosing} {Datasets} with
  {Training} {Dynamics}.
\newblock In \emph{Proceedings of the 2020 {Conference} on {Empirical}
  {Methods} in {Natural} {Language} {Processing} ({EMNLP})}, pp.\  9275--9293,
  Online, November 2020. Association for Computational Linguistics.
\newblock \doi{10.18653/v1/2020.emnlp-main.746}.
\newblock URL \url{https://aclanthology.org/2020.emnlp-main.746}.

\bibitem[Toneva et~al.(2019)Toneva, Sordoni, Combes, Trischler, Bengio, and
  Gordon]{dp_toneva_forgettability_2019}
Toneva, M., Sordoni, A., Combes, R. T.~d., Trischler, A., Bengio, Y., and
  Gordon, G.~J.
\newblock An {Empirical} {Study} of {Example} {Forgetting} {During} {Deep}
  {Neural} {Network} {Learning}, November 2019.
\newblock URL \url{http://arxiv.org/abs/1812.05159}.
\newblock arXiv:1812.05159 [cs, stat].

\bibitem[Tosh et~al.(2021)Tosh, Krishnamurthy, and Hsu]{tosh2021contrastive}
Tosh, C., Krishnamurthy, A., and Hsu, D.
\newblock Contrastive estimation reveals topic posterior information to linear
  models.
\newblock \emph{J. Mach. Learn. Res.}, 22:\penalty0 281--1, 2021.

\bibitem[Wang \& Isola(2020)Wang and Isola]{wang2020understanding}
Wang, T. and Isola, P.
\newblock Understanding contrastive representation learning through alignment
  and uniformity on the hypersphere.
\newblock In \emph{International Conference on Machine Learning}, pp.\
  9929--9939. PMLR, 2020.

\bibitem[Zbontar et~al.(2021)Zbontar, Jing, Misra, LeCun, and
  Deny]{zbontar2021barlow}
Zbontar, J., Jing, L., Misra, I., LeCun, Y., and Deny, S.
\newblock Barlow twins: Self-supervised learning via redundancy reduction.
\newblock In \emph{International Conference on Machine Learning}, pp.\
  12310--12320. PMLR, 2021.

\end{thebibliography}
\bibliographystyle{icml2023}

\newpage
\appendix
\onecolumn
\section{Extension to Experiments}
\label{appendix:experiments}

\subsection{Details for STL10 Investigatory Experiments}

\textbf{BYOL} We consider a ResNet-18 trained for 40 epochs on STL10 with batch size 64 using SGD with learning rate of $0.001$.

\subsection{Easy Examples are Important}

\begin{figure}[H]
\centering
    \subfigure[Forgetting Score]{\includegraphics[width=0.4\linewidth]{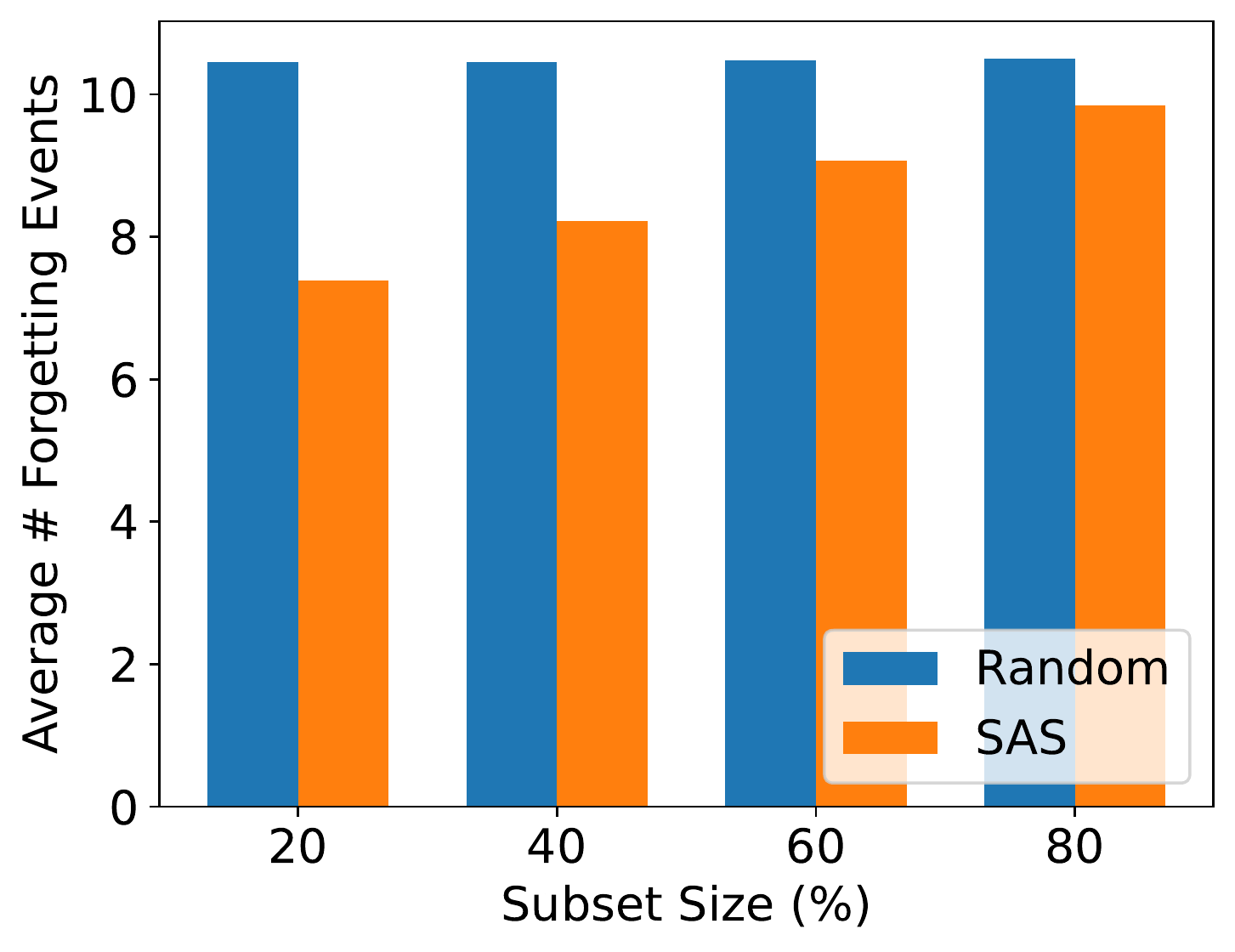} \label{fig:fscore}}
    \subfigure[Confidence]{\includegraphics[width=0.4\linewidth]{Fig/fscore.pdf}\label{fig:confidence}}
  \caption{Examples found by \method\ are easy (smaller number of forgetting events or higher confidence) for supervised learning. }
\label{fig:fscore_confidence}
\end{figure}

Here, we present results showing that the subsets \method\ selects are easier for supervised learning by various metrics. We consider the number of forgetting events \cite{dp_toneva_forgettability_2019} and
 the confidence of the prediction to quantify difficulty of a given example. Fig. \ref{fig:fscore_confidence} shows that \method\ consistently picks examples with lower average forgetting events and higher confidence than the random subsets.

\subsection{Empirical Proof of Good Alignment and Divergence}
\begin{figure}[h]
\centering
\subfigure[$\LL_{align}(S)$]{\includegraphics[width=0.4\linewidth]{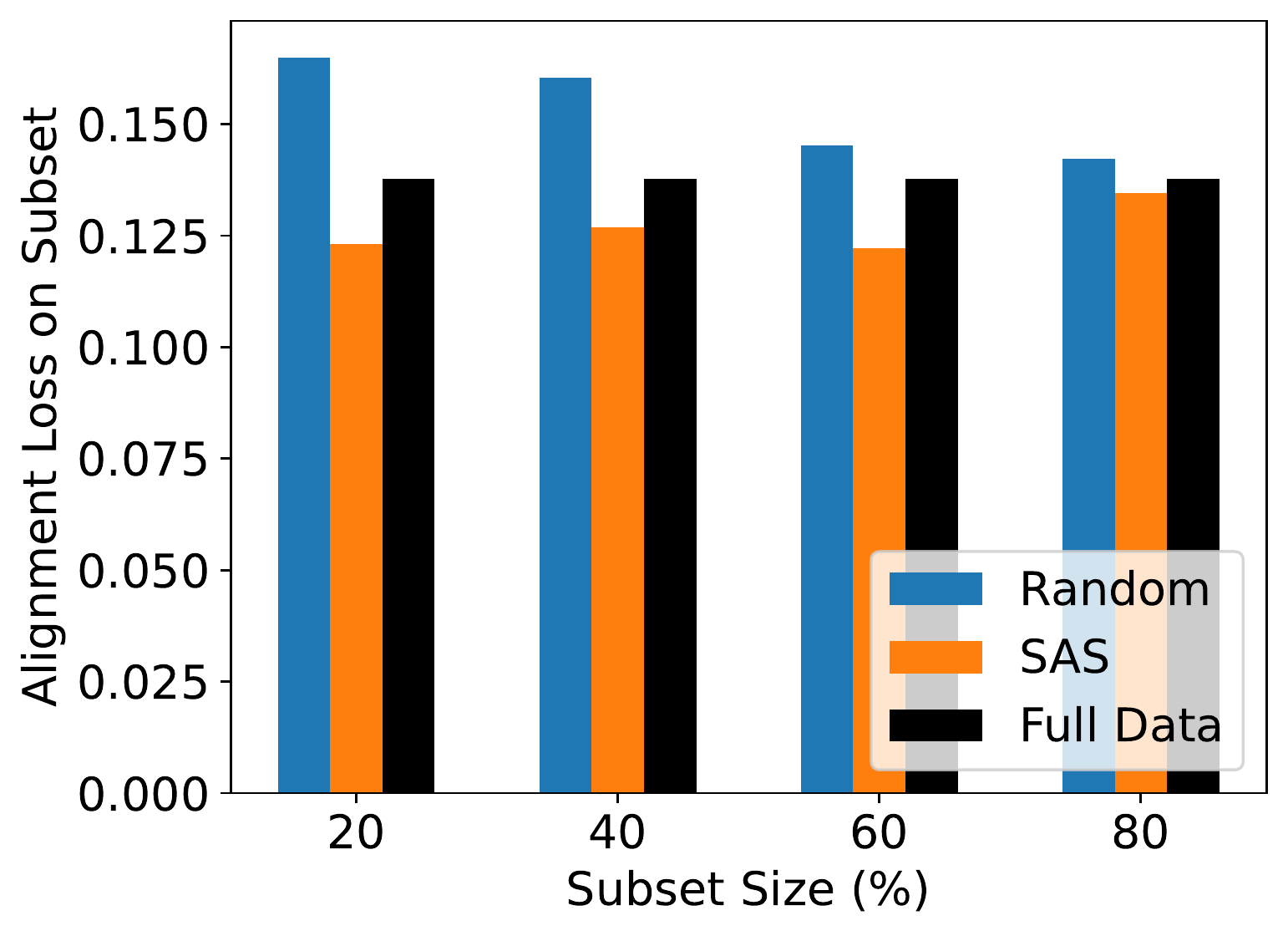} \label{fig:align}}
\subfigure[Divergence]{\includegraphics[width=0.4\linewidth]{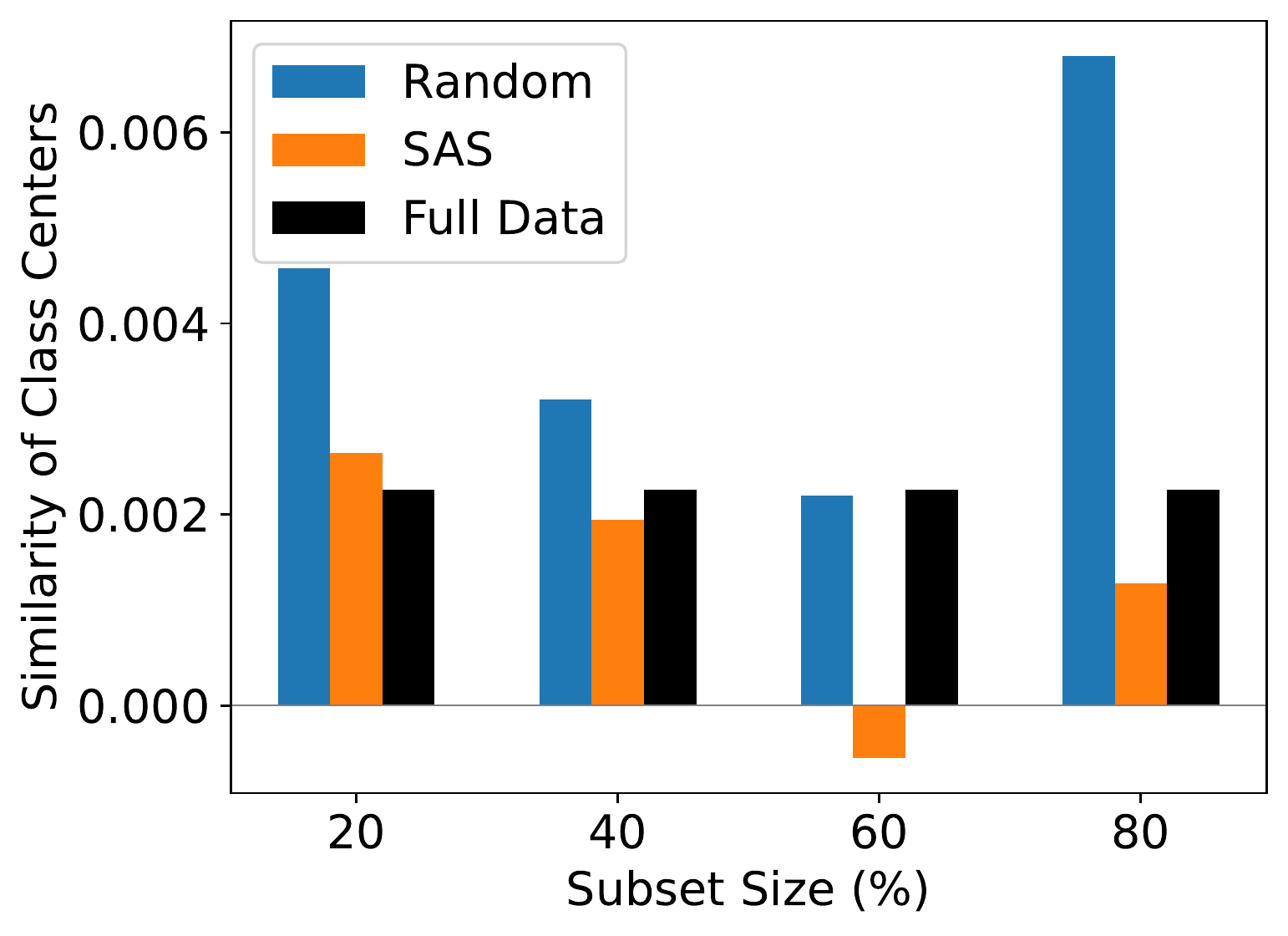} \label{fig:diverge}}
\vspace{-4mm}
\caption{Empirically verifying we find subsets that achieve good alignment and divergence} 
\label{fig:align_diverge}
\end{figure}

In Fig. \ref{fig:align_diverge}, we empirically measure $\LL_{align}(S)$ and the mean similarity of class centers to show that the subsets chosen by \method\ do indeed have better alignment and divergence than random subsets. Moreover, Fig. \ref{fig:align} 
 also empirically verifies our claim that $\LL_{align}(S_k)\leq \LL_{align}(V_k)$
\newpage

\section{Proof for Theorem \ref{thm:diverge}}
\label{appendix:proof}

\begin{proof}
First, we bound the discrepancy in divergence of subset class centers and divergence of full data class centers, relying on the discrepancy between class centers on the subset and full data. 

Let $\nu_{\mu}^k=\|\muu_k^S-\muu_k\|$ and $\nu_{\mu}^l=\|\muu_l^S-\muu_l\|$. Then,
\begin{align}
{\muu_k^S}^T {\muu_l^S} - \muu_k^T \muu_l &= {((\muu_k^S - \muu_k) + \muu_k)}^T((\muu_l^S - \muu_l) + \muu_l) - \muu_k^T \muu_l \\
&= {(\muu_k^S - \muu_k)}^T (\muu_l^S - \muu_l) + {(\muu_k^S - \muu_k)}^T \muu_l + \muu_k^T (\muu_l^S - \muu_l) + \muu_k^T \muu_l  - \muu_k^T \muu_l \\
&\leq \nu_{\mu}^k \nu_{\mu}^l + \nu_{\mu}^k \norm{\muu_l} + \nu_{\mu}^l \norm{\muu_k}
\end{align}
Thus, for a normalized encoder $\|f\|= r$ we get
\begin{align}
    {\muu_k^S}^T {\muu_l^S} - \muu_k^T \muu_l &\leq r(\nu_{\mu}^k+\nu_{\mu}^l)+\nu_{\mu}^k\nu_{\mu}^l. \label{eq:class_center_discrepancy}
\end{align}

Next, we use Theorem \ref{thm:huang_full} to provide a generalization guarantee for the downstream NN classifier.

Let $V^\epsilon \subseteq V$ be the subset of examples of the full data that are well-aligned i.e. $\forall \x_i \in V^\epsilon$, s.t. $\sup_{\x_1,\x_2\in A(\x_i)} \norm{f(\x_1)-f(\x_2)} \leq \epsilon$

Recall $V_k^0\subseteq V_k$ is the subset of examples with sharp concentration of augmented data in latent class $k$, i.e., $\sup_{i,j\in V_k^0} \min_{\x\in A(\x_j),\x'\in A(\x_j)}\norm{\x-\x'}\leq \delta$ and $|V_k^0|\geq \sigma |V_k|$ for $\sigma\in (0,1]$

\begin{theorem}[Complete version of Theorem \ref{thm:huang_main_paper} \cite{huang2021towards}]
For any $l, k \in [K]$, if 
\begin{equation}
{\muu_k}^T{\muu_l} < \phi(\sigma, \delta, \epsilon) =  r^2 (1-\rho_k (\sigma,\delta,\epsilon)-\sqrt{2\rho_k(\sigma,\delta,\epsilon)}-\frac{1}{2}\Delta_\mu),
\label{eq:lemma_3.2_huang}
\end{equation}
then every example in $V_k^0 \cap V^\epsilon$ can be classified correctly by the NN classifier, where $\rho_k(\sigma,\epsilon,\delta)=2(1-\sigma)+\frac{R_{\epsilon}}{p_k}+(\sigma-\frac{R_{\epsilon}}{p_k})(\frac{L\delta}{r}+\frac{2\epsilon}{r})$, $p_k$ = probability of an example being from latent class $k$ and $\Delta_{\mu}= 1 - \min_k\|\muu_k\|^2 / r^2$.

If for any latent class $k \in [K]$, all examples in $V_k^0 \cap V^\epsilon$can be classified correctly by the NN classifier,
then the downstream error rate of NN classifier
\begin{align}
    \xi(g_{f}(V))\leq (1-\sigma)+R_{\epsilon}(V)
\end{align}
\label{thm:huang_full}
\end{theorem}

The above Theorem cannot be directly used as the training on the subset introduces an additional error in capturing the alignment for latent class $k$, i.e., $\nu_R^k$. 
Incorporating this, we get:
\begin{align}
    {\muu_k}^T{\muu_l} < r^2 (1-\rho'_k (\sigma,\delta,\epsilon)-\sqrt{2\rho'_k(\sigma,\delta,\epsilon)}-\frac{1}{2}\Delta_\mu),
\end{align}
where $\rho'_k(\sigma,\epsilon,\delta)=2(1-\sigma)+\frac{R_{\epsilon}+\nu_R^k}{p_k}+(\sigma-\frac{R_{\epsilon}+\nu_R^k}{p_k})(\frac{L\delta}{r}+\frac{2\epsilon}{r})$, and $R_{\epsilon}$ is the probability of examples not having aligned augmented views and $\nu_R^k$ is the alignment error on latent class $k$ due to training on the subset.




From \eqref{eq:class_center_discrepancy},  we have:
\begin{align}
    {\muu_k^S}^T{\muu_l^S} + r(\nu_{\mu}^k+\nu_{\mu}^l)+\nu_{\mu}^k\nu_{\mu}^l
    &<\! r^2 \Big(1-\rho'_k(\sigma,\epsilon,\delta)
    -\sqrt{2\rho'_k(\sigma,\epsilon,\delta)}-\frac{1}{2}\Delta_\mu \Big).
\end{align}
Then, as long as the following bound on the divergence of the class centers of the subset holds:
\begin{align}
    {\muu_k^S}^T{\muu_l^S}
    &<\! r^2 \Big(1-\rho'_k(\sigma,\epsilon,\delta)
    -\sqrt{2\rho'_k(\sigma,\epsilon,\delta)}-\frac{1}{2}\Delta_\mu \Big) - r(\nu_{\mu}^k + \nu_{\mu}^l) - \nu_{\mu}^k\nu_{\mu}^l,
    \label{eq:diverge_bound}
\end{align}
by Theorem \ref{thm:huang_full}, we have that the NN classifier can correctly classify all the examples in $V_k^0 \cap V^\epsilon$ for any latent class $k \in [K]$

Thus, then incorporating our additional error in alignment $\nu_R$ into the generalization error bound in Theorem \ref{thm:huang_full}, we get
\begin{align}
    \label{eq:subset_generalization}
    \xi(g_{f^S}(V))\leq (1-\sigma)+R_{\epsilon}(V) + \nu_R 
\end{align}
\end{proof}

Now, we can bound how much smaller the inner product of the class centers on the subset must be than that on the full data to achieve equivalent generalization guarantees (Eq. \eqref{eq:subset_generalization}), i.e. how much better the divergence on the subset should be than divergence on the full data. 

Let $\varepsilon_{k,l}=r(\nu_{\mu}^k+\nu_{\mu}^l)+\nu_{\mu}^k\nu_{\mu}^l$.
Then, comparing the bounds on divergence from Eq. \eqref{eq:lemma_3.2_huang} from Theorem \ref{thm:huang_full} and Eq. \eqref{eq:diverge_bound}, we have
\begin{align}
    & r^2 (1-\rho_k(\sigma,\delta,\epsilon)-\sqrt{2\rho_k(\sigma,\delta,\epsilon)}-\frac{1}{2}\Delta_\mu)
    - r^2 (1-\rho'_k(\sigma,\delta,\epsilon)-\sqrt{2\rho'_k(\sigma,\delta,\epsilon)}-\frac{1}{2}\Delta_\mu) 
    + \varepsilon_{k,l}\\
    &=r^2\Big(\rho'_k(\sigma,\delta,\epsilon)-\rho_k(\sigma,\delta,\epsilon)+
    \sqrt{\rho'_k(\sigma,\delta,\epsilon)}-\sqrt{\rho_k(\sigma,\delta,\epsilon)})
    \Big)+\varepsilon_{k,l}.
\end{align}

Let $\zeta=\frac{\nu_R^k}{p_k}(1 - \frac{L\delta+2\epsilon}{r})$ where $p_k$ is probability of an example being from latent class $k$.

Since $\sqrt{x+a}-\sqrt{x+b}\approx\frac{a-b}{2\sqrt{x}}$ for large $x$, we get:

\begin{align}
     &\approx r^2\Big(\zeta + \frac{\zeta}{2\sqrt{\rho(\sigma,\delta,\epsilon)}}\Big) + {\nu^k_{\mu}}^2 + 2r\nu^k_{\mu} + r^2
    + r(\nu_{\mu}^k+\nu_{\mu}^l)+\nu_{\mu}^k\nu_{\mu}^l\\
    &= C \nu^k_R+2(\max\{\nu^k_{\mu},\nu_{\mu}^l\})^2+4\max\{\nu^k_{\mu},\nu_{\mu}^l\}.
    \label{eq:extra_divergence}
\end{align}
where $C=\frac{r^2}{p_k}(1-\frac{L\delta+2\epsilon}{r})(1+\frac{1}{2\sqrt{\rho_k(\sigma,\delta,\epsilon)}})$.

Hence, we can rewrite Eq. \eqref{eq:diverge_bound} as
\begin{align}
    {\muu_k^S}^T{\muu_l^S}
    &<\! \phi(\sigma, \delta, \epsilon) - \big( C \nu^k_R+2(\max\{\nu^k_{\mu},\nu_{\mu}^l\})^2+4\max\{\nu^k_{\mu},\nu_{\mu}^l\} \big)
    \label{eq:rewrite_diverge_bound}
\end{align}

When examples in every class have a high concentration of augmented data, i.e., when $\delta$ is small, $\rho(\sigma,\delta,\epsilon)$ is small and $C$ is large. However, in this settings, picking a subset according the objective in Eq. \eqref{eq:r_error} guarantees a very small $\nu_R^k$. Therefore, $C\nu_R^k$ is small. On the other hand, when examples in every class do not have a high concentration of augmented data, $\delta$ is relatively large and hence $C$ is small. As a result, $C\nu_R^k$ in Eq. \eqref{eq:rewrite_diverge_bound} is small in both cases. Thus, for small $\nu_{\mu}$, the required divergence of subset class centers for the model trained on the subset is similar to the required divergence of full data class centers for the model trained on full data.

\section{Detailed Steps to derive Eq. \eqref{eq:align_bound_2} from Eq. \eqref{eq:align_bound_1}}
\label{appendix:detailed_steps}

Let $j \in S_k$ and $\x_{j'}=\argmin_{\x_{j_k} \in A(\x_j)}\E_{\x_1 \in A(\x_i)}\norm{f(\x_1)-f(\x_{j_k})}$.

Then $\forall i \in V_k\setminus S_k$:
\begin{align}
    &\E_{\x_1,\x_2\in A(\x_i)} \norm{f(\x_1)-f(\x_2)} \\
    &\leq \E_{\x_1,\x_2\in A(\x_i)} [ \norm{f(\x_1)-f(\x_{j'})}+\norm{f(\x_{j'})-f(\x_2)}]\\
    &\leq \E_{\x_1 \in A(\x_i)} \norm{f(\x_1)-f(\x_{j'})} + \E_{\x_2 \in A(\x_i)} \norm{f(\x_{j'})-f(\x_2)}.
\end{align}
But by definition of $\x_{j'}$, we have:
\begin{align}
    &\leq \E_{\substack{\x_1 \in A(\x_i) \\ \x_{j_k} \in A(\x_j)}} \norm{f(\x_1)-f(\x_{j_k})} + \E_{\substack{\x_2 \in A(\x_i) \\ \x_{j_k} \in A(\x_j)}} \norm{f(\x_{j_k})-f(\x_2)}\\
    &= 2 \E_{\substack{\x_1 \in A(\x_i) \\ \x_2 \in A(\x_j)}} \norm{f(\x_1)-f(\x_2)}.
\end{align}

Since this inequality holds for any $j \in S$, we get:
\begin{align}
    &\E_{\x_1,\x_2\in A(\x_i)} \norm{f(\x_1)-f(\x_2)} \leq 2 \min_{j \in S} \E_{\substack{\x_1 \in A(\x_i) \\ \x_2 \in A(\x_j)}} \norm{f(\x_1)-f(\x_2)}.
\end{align}

Thus, substituting the aforementioned bound to upper bound the second term (summation over $i \in V_k \setminus S_k$) Eq. \eqref{eq:align_bound_1}, we get Eq. \eqref{eq:align_bound_2} i.e.:
\begin{align}
&\frac{\eta(\epsilon)}{n_k} \!\cdot\! \Bigg( \sum_{i \in S_k}\E_{\x_1,\x_2\in A(\x_i)} \norm{f(\x_1)-f(\x_2)} + \sum_{i \in V_k\setminus S_k}\E_{\x_1,\x_2\in A(\x_i)} \norm{f(\x_1)-f(\x_2)} \Bigg)  \\
&\leq \frac{\eta(\epsilon)}{n_k} \!\cdot\! \Bigg( \sum_{i \in S_k}\E_{\x_1,\x_2\in A(\x_i)} \norm{f(\x_1)-f(\x_2)} + \sum_{i \in V_k\setminus S_k}[2 \min_{j\in S_k}\E_{\substack{\x_1\in A(\x_i),\\\x_2\in A(\x_j)}} \norm{f(\x_1)\!-\!f(\x_2)}]\!\Bigg).
\end{align}

\end{document}